\documentclass[journal]{IEEEtran}

\usepackage{lineno,hyperref}
\usepackage{amsmath,amssymb,amsfonts}
\usepackage{algorithmic}
\usepackage{algorithm}
\usepackage{graphicx}
\usepackage{textcomp}
\usepackage{subfigure}
\usepackage[misc]{ifsym}
\usepackage{color}
\usepackage{booktabs}
\usepackage{multirow}
\usepackage{times}
\usepackage{float}
\usepackage{epsfig}
\usepackage{epstopdf}
\usepackage{hyperref}
\usepackage{mathrsfs,amsmath}
\begin{document}
%
\title{Active Learning for Deep Visual Tracking}

\author{Di Yuan,
       Xiaojun Chang,
       Yi~Yang,
       Qiao Liu,
       Dehua Wang,
       and  Zhenyu He,~\IEEEmembership{Senior Member,~IEEE} 
\thanks{D. Yuan is  with Guangzhou Institute of Technology, Xidian University, Guangzhou, 510555 China (e-mail: dyuanhit@gmail.com).}
\thanks{X. Chang and Y. Yang are with the Australian Artificial Intelligence Institute, Faculty of Engineering and Information Technology, University of Technology Sydney, Australia (e-mail: \{xiaojun.chang, yi.yang\}@uts.edu.au).}
\thanks{Q. Liu is with the National Center for Applied Mathematics, Chongqing Normal University, Chongqing, 401331 China(e-mail:liuqiao.hit@gmail.com).}
\thanks{D. Wang and Z. He (Corresponding author) are with the School of Computer Science and Technology, Harbin Institute of Technology, Shenzhen, 518055 China (e-mail:takwah@126.com; zhenyuhe@hit.edu.cn).}
}

\markboth{IEEE Transactions on Neural Networks and Learning Systems}%
{D. Yuan \MakeLowercase{\textit{et al.}}: Active Learning for Deep Visual Tracking}


\maketitle


\begin{abstract}
Convolutional neural networks (CNNs) have been successfully applied to the single target tracking task in recent years. Generally, training a deep CNN model requires numerous labeled training samples, and the number and quality of these samples directly affect the representational capability of the trained model. However, this approach is restrictive in practice, because manually labeling such a large number of training samples is time-consuming and prohibitively expensive. In this paper, we propose an active learning method for deep visual tracking, which selects and annotates the unlabeled samples to train the deep CNNs model.  Under the guidance of active learning, the tracker based on the trained deep CNNs model can achieve competitive tracking performance while reducing the labeling cost. More specifically, to ensure the diversity of selected samples, we propose an active learning method based on multi-frame collaboration to select those training samples that should be and need to be annotated. Meanwhile, considering the representativeness of these selected samples, we adopt a nearest neighbor discrimination method based on the average nearest neighbor distance to screen isolated samples and low-quality samples. Therefore, the training samples subset selected based on our method requires only a given budget to maintain the diversity and representativeness of the entire sample set. Furthermore, we adopt a Tversky loss to improve the bounding box estimation of our tracker, which can ensure that the tracker achieves more accurate target states. Extensive experimental results confirm that our active learning-based tracker (ALT) achieves competitive tracking accuracy and speed compared with state-of-the-art trackers on the seven most challenging evaluation benchmarks.
\end{abstract}
\IEEEpeerreviewmaketitle

\section{Introduction}
\IEEEPARstart{S}{ingle} target tracking is a challenging and important task in the computer vision community, with numerous applications including video surveillance, autonomous vehicles, etc.
In the tracking task, the core problem is to predict the target state in all subsequent frames when the target ground-truth is provided at the beginning \cite{OTB100,UAV123,CFTM}.
Although deep learning-based trackers have obtained some notable tracking results in recent years, it remains a difficult problem to train a deep CNNs model with strong representational ability 
due to the limitations associated with labeled training samples, as well as the time-consuming and expensive nature of annotating unlabeled samples.

Recently, trackers based on the deep CNNs models have achieved competitive tracking accuracy \cite{CFNet,ATOM,D3S,CCRT,RenXCHLCW21}.
In contrast to traditional trackers \cite{MMTV,NONT,DSTRT}, these trackers need to have a large number of labeled samples available in advance to train their deep CNNs models to track the target of interest \cite{SiamFC,DiMP,C-RPN}.
The key to determining the representational ability of the deep CNNs model lies in the diversity and comprehensiveness of the labeled training samples (sequences). However, the number of training samples that have labels is very limited,  which directly limits the tracking performance of those trackers whose models are trained on these samples.
To ensure diversity and comprehensiveness, it is necessary to collect as many training sequences as possible and label the objects in each sequence.
Labeling small amounts of training sequences is tractable, however, labeling large amounts of training sequences is time-consuming and potentially expensive \cite{TrackingNet,LaSOT,GOT10k}.
Some methods attempt to use a small amount of labeled data and a large amount of unlabeled data for model training and have achieved acceptable results \cite{SSLCN,AALS}.
Another effective solution is to randomly select a subset of a large unlabeled sequence set for labeling to train the deep CNNs models.
In the case of random selection, only the selected subset is sufficiently large to ensure the diversity and representativeness of the entire unlabeled training sample set.
The alternative approach to random selection is the use of an active learning method to select samples that can improve the diversity of the sample space \cite{ALS,gal2017deep,LiuC0Y17,RenXCHLCW20,LiuCCPZYH20,YangMNCH15,ChangNWYZZ16}.
The general hypothesis is that a model trained on the sample subset, selected based on the active learning method, typically has a stronger representation ability than a model trained on the randomly selected subset under the same budget (number of samples). 
Therefore, using an active learning method to select a subset of whole training samples for labeling can not only reduce labeling costs but also improves the representation ability of models trained on this selected and labeled sample subset \cite{aghdam2019active,ALIC,pinsler2019bayesian}.

Based on the above analysis, we propose to perform an active learning method on the single target tracking task.
The active learning method we used to select the labeling subset sequences from numerous unlabeled video sequences to train the deep CNNs model of the tracker. The proposed method can ensure that the representational capability of the trained model can be improved as much as possible when the sample sequence budget is fixed.
In this way, the tracker based on the trained deep CNNs model can achieve competitive tracking performance while reducing labeling costs. 
This differs from other tasks (such as image classification \cite{gal2017deep,ALIC,yoo2019learning}, image segmentation \cite{BSegm,ALSegm,SNNL}) in which active learning is used to select image samples, the tracking task applies the active learning method for video sequences selection.
The active learning method is only based on selecting a single frame image information in the sequence, and labeling the training sample sequences will affect the representational ability of the selected sample sequences due to the background interference.
Considering the time consistency of the moving target in each video, we present the active learning sequence selection method in a multi-frame collaboration way.
This approach can eliminate the background interference through the temporal relation between multiple frames in each video sequence, thereby ensuring the diversity of the selected sample sequences.
Meanwhile, considering the representativeness of sample sequences, we adopt a nearest neighbor discrimination method based on the threshold used to screen isolated sequences. 
Therefore, our proposed active learning method can ensure the diversity and representativeness of the selected training sample sequences using a limited budget.

\begin{figure}[!t]
\centering
\includegraphics[width=0.5\textwidth]{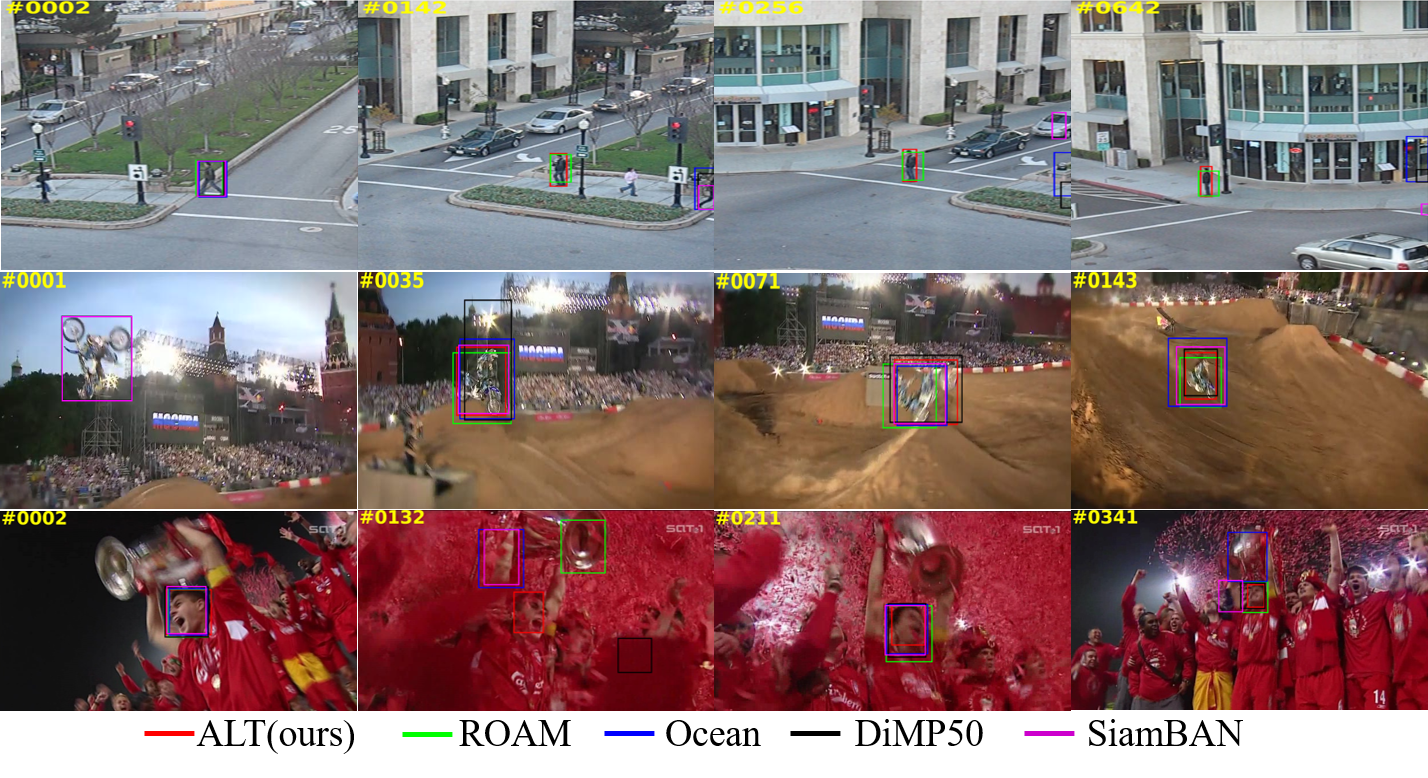}
\caption{A visual experimental comparison of the proposed ALT tracker with some state-of-the-art trackers. Thanks to the training samples selected by active learning and the target boundary box estimation improved by Tversky loss, our ALT tracker was able to obtain more accurate tracking results.}
\label{fig1}
\end{figure}

Another problem that directly affects the tracking performance is the accuracy of bounding box estimation.
Current tracking algorithms mainly use the following methods to estimate the bounding box of the target: using multi-scale factors to estimate the target state \cite{C-COT,ECO,DMLST}, or taking some proposals and using IoU scores to determine target states \cite{ATOM,DiMP}.
The former mostly uses certain fixed scale factors, which cannot accurately estimate the target boundary. 
The latter does not distinguish between the importance of the target and the background, which leads to insufficient attention being paid to the tracking target area.
To mitigate the negative impact of this problem on tracking performance, we adopt the Tversky loss function \cite{Tversky} to improve the tracker's bounding box estimation strategy, which enables it to obtain a more accurate target state.
Compared with other loss functions, the adopted Tversky loss function can ensure that the tracker pays more attention to the target than the background. As shown in Fig. \ref{fig1}, benefiting from the training samples selected by the active learning method and the target boundary box estimation improved by Tversky loss, our ALT tracker achieves more accurate target bounding boxes than the ROAM \cite{ROAM}, Ocean \cite{Ocean}, DiMP50 \cite{DiMP} and SiamBAN \cite{SiamBAN} trackers.

The main contributions of our research can be summarized as follows:
\begin{itemize}
\item We formulate a novel active learning method for training sample selection to train the deep CNNs model in a tracker. This method will select the most diverse and representative training samples under a given budget, which can greatly reduce the cost of labeling those training samples while ensuring acceptable tracking performance.
\item Considering the temporal relationship of the moving target in the video sequence, we present the active learning method using a multi-frame cooperation strategy for selecting these training samples, which can ensure the diversity of these selected samples. 
\item In addition, we adopt a nearest neighbor discrimination method based on the average nearest neighbor distance to screen isolated samples, this will ensure the representativeness of selected training samples, which can effectively ensure the robustness of the trained deep CNNs model.
\item Furthermore, we adopt a Tversky loss to improve the bounding box estimation strategy of the proposed tracker, which enables our ALT tracker to obtain a more accurate target state.
\item Extensive experimentation shows that our ALT tracker achieves more competitive results than other state-of-the-art trackers on some challenging tracking test datasets: OTB100 \cite{OTB100}, UAV123 \cite{UAV123}, TrackingNet \cite{TrackingNet}, LaSOT \cite{LaSOT}, GOT10k \cite{GOT10k}, VOT2019 \cite{VOT2019} and VOT2020 \cite{VOT2020}.
\end{itemize}

The remainder of this paper is organized as follows: Some related works are first presented in Section \ref{Rw}, after which the active learning-based training samples selection method is proposed to retain the diversity and representativeness of training samples in Section \ref{OurM}. Next, experimental results are outlined in Section \ref{Exp} to verify the effectiveness of our proposed active learning method for the tracking task. Finally, a brief conclusion of this paper is provided in Section \ref{Con}.

\section{Related Works} \label{Rw}
\subsection{Deep learning-based tracking methods}
Recently, deep learning-based trackers have attracted increasing attention.
Siamese network-based trackers treat the tracking task as a cross-correlation problem \cite{SiamFC,SiamHA,SINT,SSDCT,SiamDW,SiamFAFM,LuoCNYHZ18}.
The SiamFC \cite{SiamFC} tracker incorporates a fully convolutional Siamese network in an end-to-end manner, which demonstrates the powerful representation ability of the offline trained feature extraction network. 
The HASN \cite{SiamHA} tracker introduces an attention mechanism into the Siamese tracking framework, and further introduces a hierarchical attention Siamese network for the target tracking task.
Currently, Siamese network-based trackers enhance their tracking accuracy by adding a region proposal network \cite{SiamRPN++,SiamAttn,SiamRCNN} or some other kind of module \cite{SiamBAN,SiamCAR,ALSST}. 
All of these Siamese network-based trackers require abundant training samples to train their matching networks.
To dynamically optimize the hyperparameter in the visual tracker, Dong $et$ $al.$  \cite{DHODRL} propose adaptively optimizing hyperparameters using a deep reinforcement learning method.
The proposed method could easily be equipped with the Siamese network-based trackers and the correlation filter-based trackers for better tracking performance.
Other deep learning-based trackers have been inspired by the fast calculation in the correlation filter tracking framework, and have therefore attempted to treat the correlation filter as a network-layer and add it into the deep network to obtain a faster tracking speed \cite{CFNet,C-COT,ECO,DOTSL}.
All of these deep correlation tracking methods use an off-the-shelf feature extraction network (e.g., VGGNet or AlexNet) to fine-tune the tracking task.
In order to balance the difference between positive and negative training samples, the DSLT++ tracker \cite{DOTSL} provided a shrinkage loss to better distinguish target objects from the background by penalizing the importance of easy training data that mostly comes from the background.
The ATOM \cite{ATOM} tracker divides the tracking task into tasks of classification and estimation. 
The classification task aims to provide a rough location of the target. 
The estimation task will then predict the exact target state, dependent on the given classification information.
In \cite{DiMP}, a discriminative learning loss has been proposed that exploits target and background information for target model prediction, which can effectively improve the tracking accuracy of the tracker.
Since the confidence score lacks a clear probabilistic interpretation, the PrDiMP \cite{PrDiMP} tracker was developed to propose a probabilistic regression formulation for the tracking task.
Although these trackers described above have achieved remarkable tracking performance, they are highly dependent on the number of labeled training samples \cite{ATOM,DiMP,SiamRPN++}.
Accordingly, unlike the above deep CNNs model-based trackers that use a trained deep CNNs model with numerous manually labeled training samples, the active learning method is adopted in this paper used to select a subset of training samples to train the deep CNNs model.
The trained model is not only able to achieve considerable tracking performance but also effectively avoids the problems of being time-consuming and prohibitively expensive.

\subsection{Active learning for training samples selection}
The most important role of active learning methods is that of querying the next sample that should be labeled.
There are some existing query strategies in active learning methods: informativeness \cite{gal2017deep}, representativeness \cite{sener2018active}, hybrid \cite{huang2014active}, and performance-based \cite{fu2018scalable}.
Through the use of these query strategies, active learning methods have been applied to text classification \cite{yan2020active}, image classification \cite{gal2017deep,ALIC,yoo2019learning}, object detection \cite{roy2018deep,aghdam2019active,bengar2019temporal}, etc.
In \cite{gal2017deep}, the BALD method integrates Bayesian deep learning into an active learning framework for high-dimensional data.
In \cite{yoo2019learning}, the authors present a task-agnostic active learning method capable of training the network from a single loss prediction block. 
Other methods attempt to reduce the labeling cost with a focus on sparse subset approximation \cite{pinsler2019bayesian} or core-set selection \cite{sener2018active}.
Separate from the task of choosing image samples based on active learning described above, our main task is to select video sequence samples based on active learning.
First, most existing active learning methods use independent image samples to sort the training samples. By contrast, we use multi-frame cooperation to sort the training sequence samples according to the internal time correlation of the sequences, which is well suited for the tracking task.
Second, most existing active learning methods indiscriminately use selected samples for model training. 
However, after considering the representativeness of these selected sequence samples, we use the nearest neighbor verification method to reorder the selected sequences, which enables us to effectively exclude isolated and abnormal training samples.

\subsection{Bounding box estimation in tracking task}
Accurate target bounding box estimation is a key factor  in ensuring the tracking performance of the tracker.
There are two strategies generally adopted by most tracking methods to estimate their target boundary box. 
The first one is to use multiple fixed scale factors to determine the appropriate boundary box after the target center location is determined \cite{C-COT,ECO,GFSDCF}. 
As a representative example, in the C-COT \cite{C-COT} tracker, several fixed scale factors are used to find the most suitable target bounding box.
The other one is to select the most appropriate proposal for determining the target boundary box  after being given certain proposals \cite{ATOM,DiMP,SiamRPN++,PrDiMP}.
The ATOM \cite{ATOM} tracker determines the appropriate target bounding-box by calculating the Intersection-over-Union (IoU) score of certain proposals and the target in the reference frame.
The former primarily uses the fixed scale factor, which  is unable to accurately estimate the target boundary frame \cite{ECO,GFSDCF}. 
The latter does not distinguish the importance of intersection and union between estimated and real bounding boxes, which leads to insufficient  attention being paid to the tracking target area \cite{ATOM,SiamRPN++}.
To mitigate the negative impact of this problem on tracking performance, we adopt the Tversky loss function \cite{Tversky} to improve the tracker's bounding box estimation strategy, which enables it to obtain a more accurate target state.
Compared with other loss functions, using the Tversky loss function can make the tracker pay more attention to the overlap between the predicted area and the ground-truth, which can  in turn ensure that the tracker's prediction is more in line with the actual target state.

\section{Active learning for object tracking} \label{OurM}
We propose an active learning method to select training samples for deep CNNs model training in the tracking framework for the visual tracking task. 
First, we present an introduction of the basic deep CNNs model-based tracking framework in Section. \ref{DCTF}. We then present the active learning approach designed to select the training sample sequences that train the deep CNNs model for object tracking in Section. \ref{ALSS}. 
Furthermore, we adopt the Tversky loss to improve the bounding-box estimation of the target in Section. \ref{TBBS}. 

\subsection{Deep CNNs model based tracking framework} \label{DCTF}
We adopt an end-to-end DiMP \cite{DiMP} tracking architecture as the basic deep CNNs model-based tracking framework.
The tracking problem in this framework incorporates a target classification component and a bounding box estimation component. 
The target \textbf{classification} component contains a convolutional block that can extract features of the image patch.  
Given labeled training samples, the classification model can generate weights of the target classifier. 
Applying the obtained weights to the features extracted from a new test image patch can enable the target confidence score to be obtained.
The target classification loss can be defined as follows:
\begin{equation} \label{clssmodel1}
L(f) = \lVert r(x * f, c) \rVert^2 + \lVert \lambda f \rVert^2,
\end{equation}
where $x$ is the image patch, $f$ is the filter weights, $*$ means convolution, $x*f$ denotes target confidence scores, $c$ is the ground-truth target center, $r(x*f, c)$ denotes the residual, and $\lambda$ is a regularization parameter. 
By introducing the discriminative learning loss, Eq.(\ref{clssmodel1}) can be rewritten as follows:
\begin{equation}\label{eq:dcf1}
L_{cl} =\lVert l(x * f', z_c) \rVert^2,
\end{equation}
where $z_c$ denotes the regression label being set to a Gaussian function centered as $c$, while $f'$ indicates the weight parameters of the feature extraction network. 

The bounding box \textbf{estimation} component adopts an overlap maximization method which is proposed in the ATOM \cite{ATOM} tracker. 
Given a set of target candidate bounding boxes based on the target confidence score obtained in the target classification component, the bounding box estimation model is trained to find the maximum IoU score between these estimate bounding boxes ($B$) and the target ground-truth bounding box ($B^{gt}$). 
The IoU loss is defined as follows:
\begin{equation}
L_{ibb} = 1 - IoU,
\end{equation}
where $L_{ibb}$ is the IoU loss of the bounding box estimation, and $IoU =\frac{B \cap B^{gt}}{B \cup B^{gt}}$. Each part in the IoU  is clearly illustrated in Fig. \ref{TLOSS}. 
The full tracking framework of the DiMP \cite{DiMP} tracker is trained by combining bounding box estimation loss and target classification loss as: $L_{tot}$ $= \eta L_{cl} + L_{ibb}$. More details can be found in literature \cite{DiMP}.

\begin{figure*} [!t]
\centering
\centerline{{\includegraphics[width=0.995\textwidth]{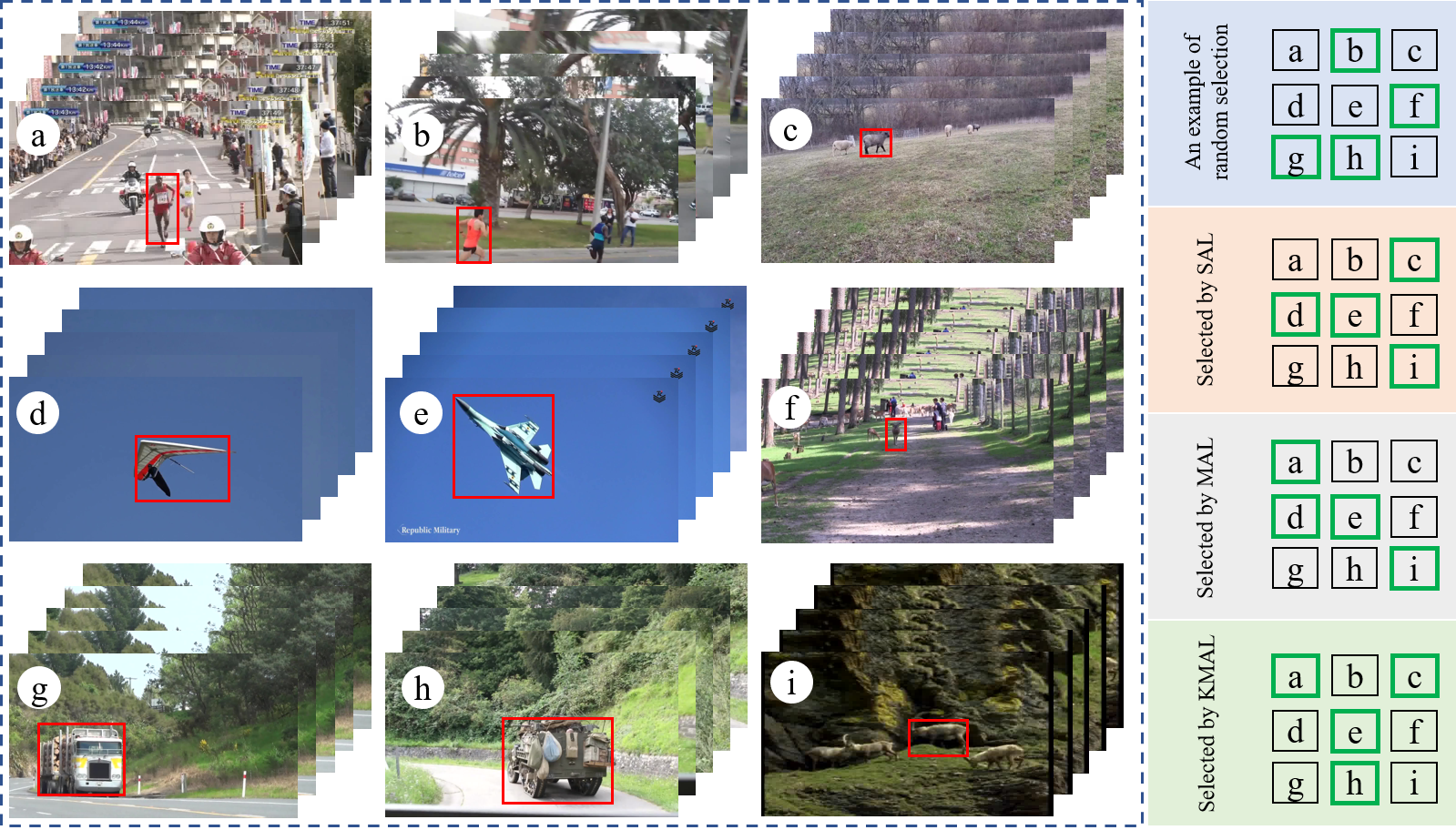}}}
\caption{An example of training sample selection under different methods: random, single frame-based active learning (SAL), multi-frame cooperation-based active learning (MAL), and neighborhood validation-based re-selection (KMAL). The bold (green) box indicates these samples selected by each of the aforementioned methods. There is a strong uncertainty in randomly selected training samples. Due to the influence of background, the training samples selected by SAL comprise only two categories: animals and aircraft. Correspondingly, the training samples selected by MAL based on multi-frame collaboration include three categories: human, animal, and aircraft. Moreover, the training samples selected by KMAL using neighborhood verification include all four categories: human, animal, vehicle, and aircraft.}
\label{fig2}
\end{figure*}

\subsection{Active learning for training sequences selection} \label{ALSS}
Owing to the limitation of training samples with labels, the CNNs models  that are trained based on these samples often lack strong representational ability, which results in the tracking performance of trackers using these models  being unable to reach the ideal state.
Although  it is easy to obtain a large number of unlabeled samples, labeling them is time-consuming and prohibitively expensive \cite{LaSOT,GOT10k,TrackingNet,ZhouCSSYN20}. 
A compromise method is to randomly select a certain number of these samples for annotation, however, this random selection method cannot guarantee the performance of the trained deep CNNs model. Consequently, a method based on active learning to select samples for annotation is proposed in this paper.
The hypothesis of active learning is to select those samples that can provide more valuable information for model training than other samples. 
Therefore, identifying an appropriate selection method is the key to selecting more representative samples.
Most of the active learning methods are based on using the similarity between image samples to select and label these selected samples.
Inspired by PointNet++ \cite{PointNet++} using the farthest point sampling (FPS) algorithm to select the input point subset to cover the whole set, we used the FPS method to make an initial selection of samples to subsequently select a representative sample subset. 
In the simplest possible terms, the FPS algorithm always selects the sample that is furthest away from all  samples in the selected subset and adds it to the subset, which ensures the diversity of samples in the selected subset. Cosine distance is used to measure the distance between two samples in this paper.
Fig. \ref{fig2} presents a simple example of training sample sequences selection. From this figure, we can clearly see that training samples selected through active learning can ensure the diversity and representativeness of samples as much as possible under the given budget  (a.k.a the number of video sequences).

\noindent \textbf{First-frame based selection method}: Given an unannotated video sequence samples set $A =\{s_1,s_2,...,s_n\}$, we use iterative farthest point sampling (FPS) method to choose a subset of sequence samples $\{s_{i_1},s_{i_2},...,s_{i_k}\}$, such that $s_{i_j}$ is the most distant sequence sample (in terms of metric distance) from the subset $\{s_{i_1},s_{i_2},...,s_{i_{j-1}}\}$ with regard to the remaining sequence samples. 
To facilitate the selection of sample sequences, the first frame image $i_{i_1}$ of each sequence $s_i$ is selected as the representative of the sequence.
In other words, given a set of images $I =\{i_{1_1},i_{2_1},...,i_{n_1}\}$, a fixed number (budget) of the image subset is selected via active learning. 
Each image $i_{i_1}$ represents a sequence $s_i$, and  the video sequence subset is labeled in correspondence to the selected image subset.
In this case, our problem pertaining to active learning for video sequence sample subset selection becomes one of image sample subset selection. 
The sample subset that is selected based on the active learning method is more diverse than the sample subset based on random selection.
Labeling these sequence samples selected based on active learning for the deep CNNs model training can improve the representational capacity of the trained model under the premise  that the budget is fixed.

\noindent \textbf{Multi-frame cooperation-based selection method}: The sequence selection method based on the first frame image may be influenced by the background information, which causes the selected sequence to pay too much attention to the background relative to the target.
However,  our goal is to learn a model that focuses more on the target than the background.
Considering the time consistency of the target in motion, we present an active learning sequence selection method based on multi-frame collaboration.
More specifically, we select multi-frame images $\{i_{i_1},i_{i_{1+a}},i_{i_{1+2a}},i_{i_{1+3a}},i_{i_{1+4a}}\}$ with the same interval $a$ in each sequence $s_i$, fuse the information of these multi-frame images selected from each sequence as $I_i$, and then select a subset $\{I_{i_1},I_{i_2},...,I_{i_k}\}$ of these fused images $\{I_1,I_2,...,I_n\}$  by using an active learning method.
The multi-frame collaboration method can eliminate the background interference through the use of temporal relation between multiple frames in each video sequence, thus ensuring the diversity of selected video sequences.

\noindent \textbf{Neighborhood validation-based re-selection method}:
Although the sequence selection method based on multi-frame validation can eliminate the influence of the background and enhance the diversity of selected sequences. Problems such as poor sequence quality and sequence isolation inevitably   emerge during the collection of unlabeled sequences. Once these poor-quality or isolated sequences are selected  for inclusion in the subset for labeling and used for model training, this will have a negative effect on the trained model.
After considering the representativeness of the selected video sequence samples, we adopt a nearest neighbor discrimination method based on the average nearest neighbor distance to screen isolated sequences.
In more detail, we compared the distance $d_i$ of the new sample $i$ with its nearest neighbor $nn_i$ and the average nearest neighbor distance $ave_{d_i}$ in each round of querying. The comparison results are used to determine whether or not the sample is added to the sample subset to be labeled.  By adopting this approach, both the diversity and the representativeness of selected samples can be guaranteed.

Algorithm \ref{alg:1} presents a general process  for a training sample selection approach based on our proposed active learning method. For the purpose of disambiguation, the video sequence sampling method mentioned here refers to the image sampling after multi-frame fusion, which corresponds to the video sequence in which it is located.
The selected subset of these samples corresponds to the selected subset of these unannotated video sequences.
The subset of samples selected based on active learning can maintain the diversity (due to the FPS algorithm) and representativeness (due to the neighborhood validation) of the original sample set   to the greatest extent possible under the fixed budget.

\begin{algorithm} [!t]
\caption{Active learning for training samples selection}
\label{alg:1}
\begin{algorithmic}[1]
\REQUIRE
An unannotated sequence samples set $A$.
\ENSURE
A diverse and representative samples subset $subA$ selected from $A$ with a given budget of $B$.
\STATE Given the distance matrix $M$ between samples in $A$;
\STATE $d_i$ is the distance between each sample $i$ in $A$ and its nearest neighbor $nn_i$, while $ave_{d_i}$ is the average nearest neighbor distance;
\STATE Randomly select a sample $i$ with $d_i \leq ave_{d_i}$ to add to the subset $subA$;
\IF{(the number of selected samples in $subA$ $<$ $B$)}
\STATE Use the FPS algorithm to select the next samples $i_s$;
  \IF{(the nearest neighbor $nn_{i_s}$ of $i_s$ is still not within $subA$ \& $d_{i_s} \leq ave_{d_i}$)}
  \STATE Add $i_s$ to the subset $subA$.
  \ENDIF
\ENDIF
\end{algorithmic}
\end{algorithm}

\subsection{Tversky loss for bounding-box estimation} \label{TBBS}
\begin{figure}[!t]
\centering
\includegraphics[width=0.475\textwidth]{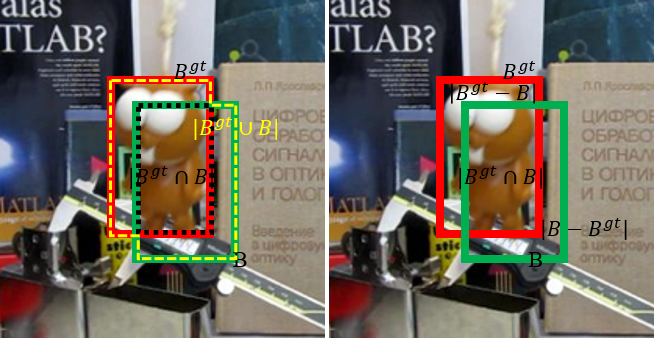}
\caption{ An example of each part in IoU loss and Tversky loss. $B$ is the predicted target bounding box, $B^{gt}$ is the target ground-truth bounding box. The black dotted box represents $|B \cap B^{gt}|$, while the yellow dotted box represents $|B \cup B^{gt}|$ in the left sub-figure. The $|B-B^{gt}|$ means treating the background as the target, and the $|B^{gt}-B|$ means treating the target as the background.}
\label{TLOSS}
\end{figure}

Another problem that directly affects the tracking performance is the accuracy of bounding-box estimation for the target. To obtain more accurate tracking results, we adopt a Tversky loss to improve the bounding box estimation strategy in the tracking framework.
The Tversky loss function can be defined according to the following format:
\begin{equation}
\centering
L_{tbb} = 1 -T(B, B^{gt}),
\end{equation}
where $L_{tbb}$ is the Tversky loss for bounding-box estimation, while $T(B, B^{gt})$ is the Tversky coefficient:
\begin{equation}
T(B, B^{gt}) = \frac{|B \cap B^{gt}|}{|B \cap B^{gt}| + \alpha |B-B^{gt}| + \beta |B^{gt}-B|}.
\end{equation}
In the Tversky coefficient $T(B, B^{gt})$, if $\alpha = \beta = 0.5$, $T(B, B^{gt})$ is the Dice coefficient, moreover, if $\alpha = \beta = 1$, $T(B, B^{gt})$ is the Jaccard coefficient. 
In $T(B, B^{gt})$, the $|B-B^{gt}|$ term means to treat the background as the target, while the $|B^{gt}-B|$ term means to treat the target as the background (shown in Fig. \ref{TLOSS}). 
 In fact, it is even harder to accept that the tracker treats the background as the target, as this can shift the target's appearance model and cause the tracker to lose its target. The Tversky loss function provides adjustable parameters used to adjust the weights of the $|B^{gt}-B|$ and $|B-B^{gt}|$  parts. Compared with the IoU loss, we can control the attention of the model by adjusting $\alpha$ and $\beta$ in the Tversky loss function, which can cause the model to pay more attention to the target than the background and avoid treating the background as the target, enabling it to obtain a more accurate boundary box of the tracking target.
We train our full tracking framework by minimizing the total loss, which is the combination of this Tversky loss with the target classification loss:
\begin{equation}
\centering
 L_{tot} = L_{tbb} + \eta L_{cl},
\end{equation}
where $\eta$ is the target classification weight used to adjust the specific gravity between the two-loss parts.

\begin{table*}[!t]
\renewcommand\arraystretch{1.5}
\scriptsize
\centering
\caption{Ablation study results on LaSOT \cite{LaSOT} benchmark datasets. 
The best scores are highlighted in {\color{red}\text{red}}.
}
\begin{tabular}{l|cc|cc|cc|cc|cc}
\toprule[1pt]
Budget & \multicolumn{2}{c|}{50} &\multicolumn{2}{c|}{100} & \multicolumn{2}{c|}{500}  &\multicolumn{2}{c|}{1000} & \multicolumn{2}{c}{2000} \\
\midrule
index & Norm. Precision & AUC & Norm. Precision & AUC & Norm. Precision & AUC & Norm. Precision & AUC & Norm. Precision & AUC\\
\midrule
Random     & $49.9$ & $42.5$ & $54.3$ & $46.5$ & $57.6$ & $51.5$ & $58.6$ & $52.6$ & $59.2$ & $53.0$ \\
SAL        & $50.2$ & $43.1$ & $55.0$ & $47.1$ & $59.4$ & $52.6$ & $59.4$ & $53.1$ & $60.8$ & $54.2$ \\
MAL        & $52.3$ & $43.9$ & $55.4$ & $47.9$ & $60.2$ & $52.9$ & $60.7$ & $53.9$ & $61.9$ & $55.4$ \\
KMAL       & $52.5$ & $44.6$ & $55.9$ & $48.3$ & $61.1$ & $53.6$ & $62.9$ & $55.6$ & $65.3$ & $57.1$ \\
TKMAL(ALT) & $\color{red}{53.7}$ &  $\color{red}{45.4}$ &  $\color{red}{58.5}$ &  $\color{red}{52.0}$ &  $\color{red}{61.9}$ &  $\color{red}{54.3}$ &  $\color{red}{63.6}$ &  $\color{red}{56.3}$ &  $\color{red}{66.3}$ &  $\color{red}{57.9}$ \\
\bottomrule[1pt]
\end{tabular}\label{table:ab}
\end{table*}

\section{Experiments} \label{Exp}
In this section, we first introduce some experimental details of our ALT tracking method. 
We next report some ablation studies to verify the effectiveness of each active learning rule for training sample selection and the Tversky loss for target bounding box regression. 
 Finally, we verified the effectiveness and competitiveness of our ALT tracker trained with limited budget samples through comparison with state-of-the-art trackers trained with extensive samples on seven standard benchmark datasets: OTB100 \cite{OTB100}, UAV123 \cite{UAV123}, GOT10k \cite{GOT10k}, LaSOT \cite{LaSOT}, TrackingNet \cite{TrackingNet}, VOT2019 \cite{VOT2019} and VOT2020 \cite{VOT2020}.

\subsection{Experimental Details}\label{ImD}
We follow the DiMP \cite{DiMP} tracker, which applies the SGD with a momentum of $0.9$ to train the deep CNNs model. 
We use ResNet-50 as our backbone network in this paper.
The weight decay is $5e$-$4$, the learning rate is $1e$-$5$, and the target classification weight $\eta$ is $10^2$.
The network is trained for $50$ epochs with a mini-batch size of $32$.
Different from the DiMP \cite{DiMP} tracker, which uses training splits of the TrackingNet \cite{TrackingNet}, LaSOT \cite{LaSOT}, GOT10k \cite{GOT10k} and COCO \cite{COCO} datasets, we  use only those training samples selected from the GOT10k \cite{GOT10k} dataset under a different budget.
In the multi-frame cooperation part, the interval $a$ is set to $10$.
In the Tversky loss part, $\alpha$ and $\beta$ are set to $0.4$ and $0.6$, respectively.
In the ablation study part, we incrementally change the training sample selection rules under the same budget to verify the effectiveness of each active learning rule. 
When the active learning rule is fixed, we gradually increase the budget value to verify that the performance of the trained network will be significantly improved  as the number of training samples increases. 
Meanwhile, the Tversky loss was added under different training sample budgets to verify its effectiveness for target bounding box estimation.
In the state-of-the-art comparison part, we used the deep CNNs model trained with the Tversky loss and with a budget of $2000$ to evaluate the tracking performance of our ALT tracker.
Our experiments are performed in Python3.7 with PyTorch on a PC with an NVIDIA GTX 2080Ti GPU. 
We run our tracker three times on each dataset and take the average tracking performance to mitigate randomness.
The average tracking speed of our ALT tracker is around $45$ fps.

\subsection{Ablation Study}
\begin{figure}[!t]
\centering
\includegraphics[width=0.5\textwidth]{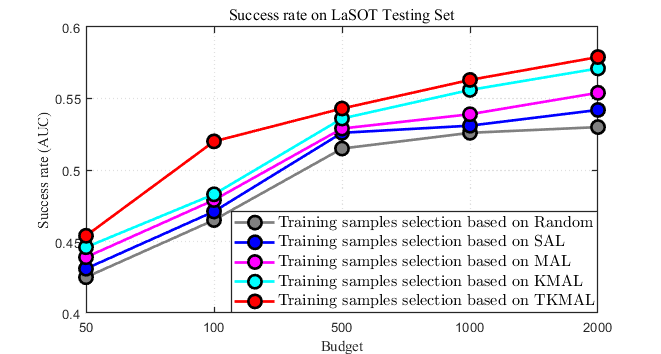}
\caption{Success rate (AUC) on different budget.}
\label{abslasot}
\end{figure}

We use the subset selected via active learning on the GOT10k \cite{GOT10k} dataset to train the deep CNNs model and carry out ablation studies on the LaSOT \cite{LaSOT} benchmark to analyze the effect of different budgets and each active learning rule under the same budget in the training process.
The comparison results are presented in Table \ref{table:ab}.
To visually demonstrate the effect of each active learning rule in the tracker, we incrementally change the  training sample selection rules. 
Starting with the basics of selecting unlabeled sequence samples based on single frame image (SAL), then changing to multi-frame coordination (MAL) based samples selection, the unlabeled training samples are re-selected by using the nearest neighbor verification (KMAL), while the target boundary box precision is improved by using the Tversky loss (TKMAL). 
Meanwhile, we assess the tracking results of the deep CNNs model trained on the random selection (Random) of the same number of samples.
As shown in Table \ref{table:ab} and Fig. \ref{abslasot}, with the same budget,  the tracking performance of the trained model exhibits increasing improvement as the sample selection rules become increasingly perfect.
Tracking performance improves with a budget under the same sample selection rules. 
 At the same time, we can see that the introduction of the Tversky loss can effectively improve tracking performance in each given budget.

\subsection{State-of-the-art comparison} 
In this section, we present some quantitative comparisons of the proposed ALT tracker with some state-of-the-art trackers on  the seven most challenging benchmark datasets to verify the effectiveness of our tracker.

\begin{table*}
\renewcommand\arraystretch{1.5}
\renewcommand\tabcolsep{4.5pt}
\scriptsize
\centering
\caption{Comparison results on OTB100 \cite{OTB100} dataset.
The top-3 scores are highlighted in {\color{red}\text{red}}, {\color{blue}\text{blue}} and {\color{green}\text{green}}, respectively.}
\label{tab:OTB100}
\begin{tabular}{l|cccccccccccc}
\toprule[1pt]
Trackers & ALT & ATOM\cite{ATOM} & GradNet\cite{GradNet} & GCT\cite{GCT} & ARCF\cite{ARCF} & UDT\cite{UDT} & DiMP50\cite{DiMP} & ROAM\cite{ROAM} &   SiamRCNN\cite{SiamRCNN} & DCFST\cite{DCFST} & PGNet\cite{PGNet}\\
\midrule
Reference & Ours & CVPR19' & ICCV19' & CVPR19' & ICCV19' & CVPR19' & ICCV19' & CVPR20'  & CVPR20' & ECCV20' & ECCV20' \\
\midrule
Precision   & $\textcolor{red}{90.8}$ & $86.2$ & $86.1$ & $85.9$ & $81.8$ & $76.0$ &  $\textcolor{green}{89.9}$ & $\textcolor{red}{90.8}$ & $89.1$ & $87.2$ & $89.2$\\
Success(AUC) & $\textcolor{green}{69.2}$ & $66.1$ & $63.9$ & $64.8$ & $61.7$ & $59.4$ & $68.7$ & $68.1$ & $ \textcolor{blue}{70.1}$ & $\textcolor{red}{70.9}$ & $69.1$\\
\bottomrule[1pt]
\end{tabular}
\end{table*}

\begin{table*}
\renewcommand\arraystretch{1.5}
\renewcommand\tabcolsep{2pt}
\scriptsize
\centering
\caption{Comparison results on UAV123 \cite{UAV123} dataset.
The top-3 scores are highlighted in {\color{red}\text{red}}, {\color{blue}\text{blue}} and {\color{green}\text{green}}, respectively.}
\label{tab:UAV123}
\begin{tabular}{l|cccccccccccc}
\toprule[1pt]
Trackers & ALT & GFSDCF\cite{GFSDCF} & DiMP50\cite{DiMP} & ARCF\cite{ARCF} & SiamRPN++\cite{SiamRPN++} & ATOM\cite{ATOM} & GCT\cite{GCT} & CGACD\cite{CGACD} & SiamBAN\cite{SiamBAN} & SiamCAR\cite{SiamCAR} & SiamAttn\cite{SiamAttn} & CLNet\cite{CLNet} \\
\midrule
Reference  & Ours & ICCV19' & ICCV19' & ICCV19' & CVPR19' & CVPR19' & CVPR19' & CVPR20' & CVPR20' & CVPR20' & CVPR20' & ECCV20'\\
\midrule
Precision    & $\textcolor{red}{87.1}$ & $76.7$ & $ \textcolor{blue}{85.8}$ & $67.6$ & $80.7$ & $\textcolor{green}{85.6}$ & $73.2$ & $83.3$ & $83.3$ & $76.0$ & $84.5$ & $83.0$\\
Success(AUC)  & $\textcolor{red}{65.2}$ & $53.4$ & $\textcolor{green}{64.8}$ & $47.0$ & $61.3$ & $64.3$ & $50.8$ & $63.3$ & $63.1$ & $61.4$ & $ \textcolor{blue}{65.0}$ & $63.3$\\
\bottomrule[1pt]
\end{tabular}
\end{table*}

\noindent \textbf{OTB100}. The OTB100 \cite{OTB100} benchmark contains two tracking performance evaluation criteria: precision score and success score. 
To test the proposed ALT tracker, we draw some experimental comparisons with some state-of-the-art trackers (namely ATOM \cite{ATOM}, GradNet \cite{GradNet}, GCT\cite{GCT}, ARCF \cite{ARCF}, UDT \cite{UDT}, DiMP50 \cite{DiMP}, ROAM \cite{ROAM}, SiamRCNN \cite{SiamRCNN}, DCFST \cite{DCFST} and PGNet \cite{PGNet}) on this dataset.
Table \ref{tab:OTB100} presents experimental results of these comparisons over all $100$ testing videos. 
It can be seen that our ALT tracker, as well as the ROAM \cite{ROAM} tracker, achieved the best precision scores. The ROAM \cite{ROAM} is trained on large-scale datasets, which is highly time consuming and expensive.
By contrast, our ALT tracker only  requires a limited number of training samples to achieve similar tracking results.
Meanwhile, by using the same backbone network (ResNet-50) as the DiMP50 \cite{DiMP} tracker, our tracker obtains some better tracking results than the DiMP50 tracker under the condition  that limited training samples are used, this fully demonstrates the effectiveness of our training sample selection strategy based on the active learning method and the accurate bounding-box estimation improved by the Tversky loss. 

\noindent \textbf{UAV123}. To evaluate the tracking performance of the proposed ALT tracker, we report some experimental comparisons of between this tracker and other state-of-the-art trackers (namely GFSDCF \cite{GFSDCF}, DiMP50 \cite{DiMP}, ARCF \cite{ARCF}, SiamRPN++ \cite{SiamRPN++}, ATOM \cite{ATOM}, GCT \cite{GCT}, CGACD \cite{CGACD}, SiamBAN \cite{SiamBAN}, SiamCAR \cite{SiamCAR}, SiamAttn \cite{SiamAttn} and CLNet \cite{CLNet}) on this UAV123 \cite{UAV123} dataset. 
Table \ref{tab:UAV123} presents the precision and success scores on $123$ video sequences. 
From this table, we can  determine that the proposed ALT tracker achieved the best tracking accuracy in terms of both the precision and success index. 
Compared to other tracking methods, the DiMP50 \cite{DiMP} tracker achieves superior tracking performance with regard to AUC ($64.8\%$) and precision ($85.8\%$) indexes.
However, the proposed ALT tracker, which uses only limited training samples and employs the Tversky loss-based bounding box estimation method,  produces a certain degree of performance improvement.
These experimental results demonstrate the effectiveness of our training sample selection method based on active learning and the improved bounding box estimation method.

\begin{figure*}[!t]
\centering
\includegraphics[width=0.99\textwidth]{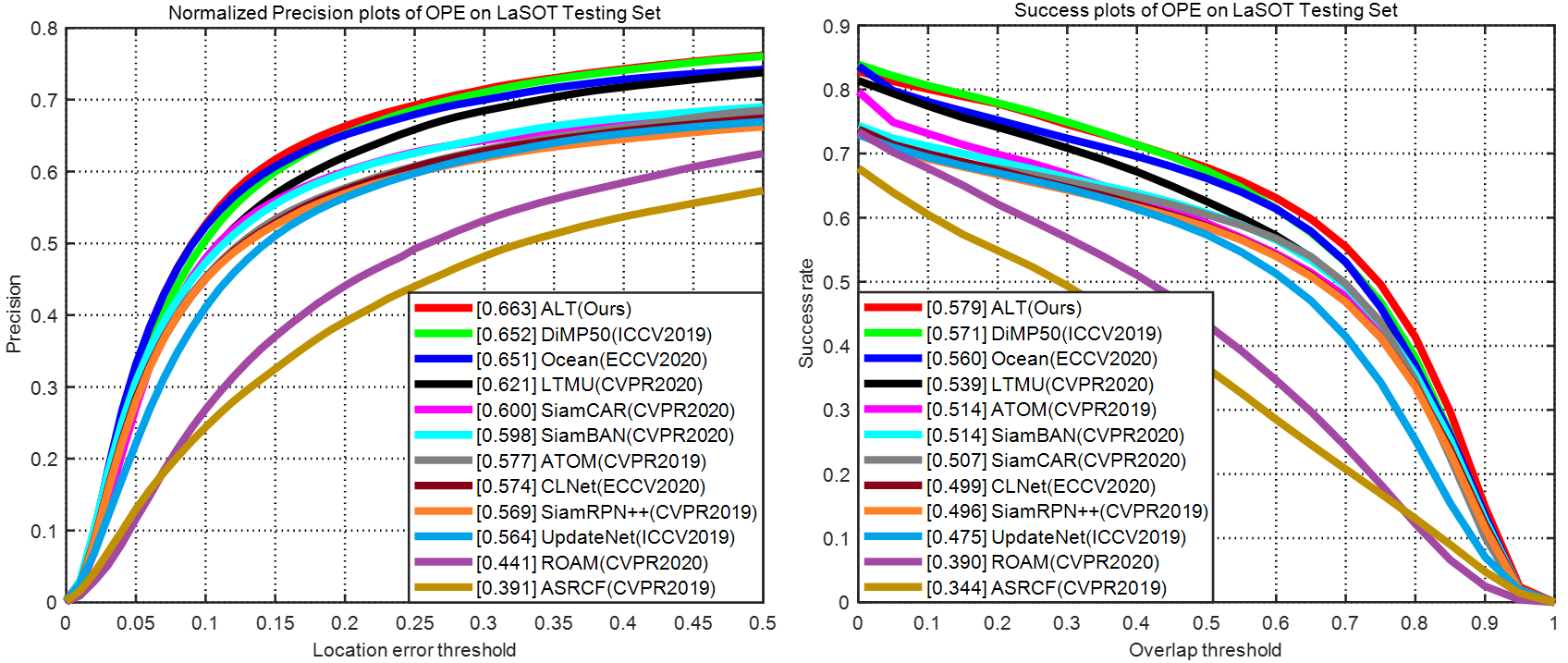}
\caption{Comparison results on LaSOT \cite{LaSOT} testing set in terms of normalized precision and success (AUC) scores.}
\label{lasot}
\end{figure*}

\noindent \textbf{GOT10k}. 
We conduct some experimental comparisons on the GOT10k \cite{GOT10k} test set to evaluate our ALT tracker in relation to other state-of-the-art trackers, namely LDES \cite{LDES}, SiamDW \cite{SiamDW}, SPM \cite{SPM}, ATOM \cite{ATOM}, DiMP18 \cite{DiMP}, DiMP50 \cite{DiMP}, SiamCAR \cite{SiamCAR}, D3S \cite{D3S}, ROAM \cite{ROAM}, Ocean \cite{Ocean}, DCFST18 \cite{DCFST} and DCFST50 \cite{DCFST}. 
The evaluation indexes include average overlap (AO) and success rate (SR$_{0.50}$, SR$_{0.75}$). 
We further show the tracking speed of each tracker.
The comparative experimental results are presented in Table \ref{tab:GOT10k}.
Although our ALT tracker requires only limited samples to train the same network structure as the DiMP50 \cite{DiMP} tracker, the tracking performance of our tracker is superior to that of the DiMP50 \cite{DiMP} tracker, as well as achieving faster tracking speed.
Meanwhile, our tracker achieved the best score on the SR$_{0.75}$ index.
The tracking accuracy of our ALT tracker is slightly lower than that of DCFST50 \cite{DCFST} on the indexes of AO and SR$_{0.50}$, however, our tracker obviously exceeds this in tracking speed.
 These comparative results show that our tracker trained with only limited samples and equipped with a boundary box estimation strategy based on Tversky loss can still obtain tracking results comparable to those of state-of-the-art trackers.

\begin{table} 
\renewcommand\arraystretch{1.5}
\renewcommand\tabcolsep{7.5pt}
\scriptsize
\centering
 \caption{Comparison results on GOT10k \cite{GOT10k} test set. The top-3 scores are highlighted in {\color{red}\text{red}}, {\color{blue}\text{blue}} and {\color{green}\text{green}}, respectively.}
 \label{tab:GOT10k}
 \begin{tabular}{l|c|cccc}
  \toprule[1pt]
  Trackers                 & Reference  & AO     & SR$_{0.50}$ &SR$_{0.75}$ & Speed \\
   \midrule
    LDES\cite{LDES}        & AAAI19'    & $35.9$ & $36.8$ & $15.3$   & $1.2$\\
    SiamDW\cite{SiamDW}    & CVPR19'    & $41.1$ & $45.6$ & $15.4$   & $12.0$\\
    SPM\cite{SPM}          & CVPR19'    & $51.3$ & $59.3$ & $35.9$   & $\color{red}\text{72.3}$\\
    ATOM\cite{ATOM}        & CVPR19'    & $55.6$ & $63.4$ & $40.2$   & $20.7$\\
    DiMP18\cite{DiMP}      & ICCV19'    & $57.9$ & $67.2$ & $44.6$   & $\color{blue}\text{57.0}$ \\
    DiMP50\cite{DiMP}      & ICCV19'    & $\color{green}\text{61.1}$ & $71.7$ & $49.2$   & $43.0$ \\
    SiamCAR\cite{SiamCAR}  & CVPR20'    & $56.9$ & $67.0$ & $41.5$   & $\color{green}\text{52.3}$ \\
    D3S\cite{D3S}          & CVPR20'    & $59.7$ & $67.6$ & $46.2$   & $25.0$ \\
    ROAM\cite{ROAM}        & CVPR20'    & $43.6$ & $46.6$ & $16.4$   & $13.0$ \\
    Ocean\cite{Ocean}      & ECCV20'    & $\color{green}\text{61.1}$ & $\color{green}\text{72.1}$ & $\color{green}\text{47.3}$   & $44.2$ \\
    DCFST18\cite{DCFST}    & ECCV20'    & $61.0$  & $71.6$	& $46.3$ & $35.0$ \\
    DCFST50\cite{DCFST}    & ECCV20'    & $\color{red}\text{63.8}$  & $\color{red}\text{75.3}$	& $\color{blue}\text{49.8}$ & $25.0$ \\
    ALT                    & Ours       & $\color{blue}\text{62.2}$ & $\color{blue}\text{73.2}$ & $\color{red}\text{50.1}$   & $44.5$ \\
    \bottomrule[1pt]
  \end{tabular}
\end{table}

\begin{table*} [http]
\renewcommand\arraystretch{1.5}
\renewcommand\tabcolsep{4.0pt}
\scriptsize
\centering
\caption{Comparison results on TrackingNet \cite{TrackingNet} test set.
The top-3 scores are highlighted in {\color{red}\text{red}}, {\color{blue}\text{blue}} and {\color{green}\text{green}}, respectively.}
\label{tab:TrackingNet}
\begin{tabular}{l|ccccccccccc}
\toprule[1pt]
Trackers & ALT & ATOM\cite{ATOM} & SPM \cite{SPM} & C-RPN\cite{C-RPN} & GFSDCF\cite{GFSDCF} & UpdateNet\cite{UpdateNet} & DiMP50\cite{DiMP}  & CGACD\cite{CGACD} & MAMLR\cite{MAML} & D3S\cite{D3S} & ROAM\cite{ROAM}\\
\midrule
Reference & Ours & CVPR19' & CVPR19' & CVPR19' & ICCV19' & ICCV19' & ICCV19' & CVPR20' & CVPR20' & CVPR20' & CVPR20' \\
\midrule
Precision   & $\color{green}\text{68.0}$ & $64.8$  & $66.1$ & $61.9$ & $56.6$  & $62.5$  & $\color{blue}\text{68.7}$    & $\textcolor{red}{69.3}$ & $48.0$  & $66.4$ & $54.7$ \\
Norm. Prec. & $\color{green}\text{79.3}$ & $77.1$  & $77.8$ & $74.6$ & $71.8$   & $75.2$  & $\color{red}\text{80.1}$     & $\color{blue}\text{80.0}$ & $78.6$  & $76.8$ & $69.5$ \\
Success(AUC)  & \color{blue}\text{73.6} & $70.3$  & $71.2$ & $66.9$  & $60.9$ & $67.7$  &   $\color{red}\text{74.0}$  & $71.1$ & $69.8$  & $\color{green}\text{72.8}$ & $62.0$ \\
\bottomrule[1pt]
\end{tabular}
\end{table*}

\noindent \textbf{LaSOT}.
To validate the tracking accuracy of our ALT tracker, we conduct several experimental comparisons on the LaSOT \cite{LaSOT} testing set with some state-of-the-art tracking methods, namely SiamRPN++ \cite{SiamRPN++}, ATOM \cite{ATOM}, ASRCF \cite{ASRCF}, UpdateNet \cite{UpdateNet}, DiMP50 \cite{DiMP}, SiamBAN \cite{SiamBAN}, SiamCAR \cite{SiamCAR}, ROAM \cite{ROAM}, LTMU \cite{LTMU}, Ocean \cite{Ocean} and CLNet \cite{CLNet}. 
Fig. \ref{lasot} presents the results of comparison on this dataset. 
Among these compared state-of-the-art trackers, the proposed ALT tracker obtains the best normalized precision and success scores. By contrast, our ALT outperforms the DiMP50 \cite{DiMP} tracker, the Ocean \cite{Ocean} tracker and the SiamCAR \cite{SiamCAR} tracker on each performance metric item, which fully proves the effectiveness of our tracking method.
Although our tracker only requires a small number of training samples, it still achieves better tracking results than DiMP50 \cite{DiMP}, Ocean \cite{Ocean} and SiamCAR \cite{SiamCAR} trackers trained on large-scale datasets, this proves that our training samples, selected based on the active learning method, have good diversity and representation.

\begin{table*}
\renewcommand\arraystretch{1.5}
\renewcommand\tabcolsep{2pt}
\scriptsize
\centering
  \caption{Comparison results on VOT2019 \cite{VOT2019} dataset. The top-3 scores are highlighted in {\color{red}\text{red}}, {\color{blue}\text{blue}} and {\color{green}\text{green}}, respectively.
  }
  \label{tab:vot2019}
  \begin{tabular}{l|cccccccccccc}
    \toprule
{Trackers} &  ATOM\cite{ATOM} & DiMP50\cite{DiMP} & PrDiMP50\cite{PrDiMP} & TADT\cite{TADT}  & SiamRPN++\cite{SiamRPN++} & MemDTC\cite{MemDTC} & Ocean\cite{Ocean} & CLNet\cite{CLNet}  & SiamBAN\cite{SiamBAN} & SiamMask\cite{SiamMask} & SPM\cite{SPM}  & ALT\\
\midrule
Reference & CVPR19' & ICCV19' & CVPR20' & CVPR19' & CVPR19' & ECCV18' & ECCV20'  & ECCV20'& CVPR20' & CVPR19'  & CVPR19' & Ours \\
\midrule
EAO ($\uparrow$)  & 0.299  &   \color{blue}\text{0.368} & 0.268 & 0.207  & 0.285  & 0.228 &  \color{green}\text{0.327} & 0.313 & \color{green}\text{0.327} & 0.287 & 0.275 & \color{red}\text{0.378}\\
Robustness ($\downarrow$) & 0.411 &  \color{red}\text{0.278} & \color{green}\text{0.355} & 0.677  & 0.482 & 0.587 & 0.376 & 0.461 & 0.393 & 0.461 & 0.507 & \color{blue}\text{0.281}\\
Accuracy ($\uparrow$)    & \color{red}\text{0.606} & 0.597  & 0.572 & 0.516  & 0.599 & 0.485  & 0.590 & \color{red}\text{0.606} & \color{green}\text{0.602}  & 0.594 & 0.577 & 0.568\\
    \bottomrule
  \end{tabular}
\end{table*}

\begin{table*}
\renewcommand\arraystretch{1.5}
\renewcommand\tabcolsep{3pt}
\scriptsize
\centering
  \caption{Comparison results on VOT2020  \cite{VOT2020} dataset. The top-3 scores are highlighted in {\color{red}\text{red}}, {\color{blue}\text{blue}} and {\color{green}\text{green}}, respectively.
  }
  \label{tab:vot2020}
  \begin{tabular}{l|ccccccccccc}
    \toprule
{Trackers} &  ATOM\cite{ATOM} & DiMP50\cite{DiMP} & A3CTDmask\cite{VOT2020} & TRAT\cite{VOT2020}  & UPDT\cite{UPDT} & DCDA\cite{VOT2020} & TCLCF\cite{VOT2020} & FSC2F\cite{VOT2020}   & CSRDCF\cite{CSRDCF} & SiamFC\cite{SiamFC}  & ALT\\
\midrule
Reference & CVPR19' & ICCV19' & ECCVW20' & ECCVW20' & ECCV18' & ECCVW20' & ECCVW20' & ECCVW20' & CVPR17'  & ECCVW16' & Ours \\
\midrule
EAO ($\uparrow$)  & 0.237  &  0.241 & \color{blue}\text{0.260} & \color{green}\text{0.256}  & 0.237  & 0.232 &0.202 & 0.156  & 0.193 & 0.172 & \color{red}\text{0.269}\\
Robustness ($\downarrow$) & 0.687 & 0.700  & \color{green}\text{0.498} & 0.724  & 0.688 & 0.624 &  0.582 &  \color{red}\text{0.465} & 0.580  & \color{blue}\text{0.479} & 0.757\\
Accuracy ($\uparrow$)    &0.440 & 0.434  &  \color{red}\text{0.634} & \color{green}\text{0.445}  & 0.443 & \color{blue}\text{0.456}  & 0.430 & 0.397  & 0.405 & 0.422 & 0.426\\
    \bottomrule
  \end{tabular}
\end{table*}

\noindent \textbf{TrackingNet}.
In order to evaluate the proposed ALT tracker, we conduct some comparisons with several state-of-the-art trackers, namely ATOM \cite{ATOM}, SPM \cite{SPM}, D3S \cite{D3S}, DiMP50 \cite{DiMP},  C-RPN \cite{C-RPN}, ROAM \cite{ROAM}, GFSDCF \cite{GFSDCF}, CGACD \cite{CGACD}, MAMLR \cite{MAML} and UpdateNet \cite{UpdateNet} on the TrackingNet \cite{TrackingNet} test set.
Table \ref{tab:TrackingNet} presents the comparison results in precision scores, normalized precision scores, and success scores. As can be seen from this table, while our tracker uses only a small number of training samples, it can achieve tracking performance resembling that achieved by these trackers when trained on a large-scale dataset.
More specifically, in the success part, our ALT outperforms the third-best tracker, compared to the best tracker (DiMP50 \cite{DiMP}), the proposed ALT tracker exhibited a decline in tracking performance of less than $0.5\%$ when the training sample was far below it. This fully demonstrates the representational ability of our training samples selected based on active learning. Compared with the IoU-based ATOM \cite{ATOM} tracker, our ALT achieves an improvement of more than $2\%$ on each metric. 
All of these comparative results show that both the active learning-based training samples selection method and the Tversky loss based bounding box estimation strategy are effective and worth practicing.

\noindent \textbf{VOT2019.} The VOT2019 dataset contains $60$ test video sequences, and trackers are evaluated using expected average overlap (EAO), robustness and accuracy.
We show some experimental comparisons of our ALT tracker with the ATOM \cite{ATOM}, DiMP50 \cite{DiMP}, PrDiMP50 \cite{PrDiMP}, TADT \cite{TADT}, SiamRPN++ \cite{SiamRPN++}, MemDTC \cite{MemDTC}, Ocean \cite{Ocean}, CLNet \cite{CLNet}, SiamBAN \cite{SiamBAN}, SiamMask \cite{SiamMask} and  SPM \cite{SPM} trackers on this dataset.
The comparison results are listed in Table \ref{tab:vot2019}.  From this table, we can know that our ALT tracker achieves the best EAO score compared to other trackers. Our ALT tracker adopts the same backbone network as the DiMP50 \cite{DiMP}, PrDiMP50 \cite{PrDiMP} and SiamRPN++ \cite{SiamRPN++} trackers, and the EAO score is significantly higher than these trackers, this indicates that our proposed method can yield more accurate tracking results.

\noindent  \textbf{VOT2020.} VOT2020 has the same dataset size as VOT2019, and trackers are also evaluated using the expected average overlap (EAO), robustness and accuracy.
We conduct some comparisons of our tracker with several state-of-the-art trackers, namely ATOM \cite{ATOM}, DiMP50 \cite{DiMP}, A3CTDmask\cite{VOT2020}, TRAT\cite{VOT2020}, UPDT\cite{UPDT}, DCDA\cite{VOT2020}, TCLCF\cite{VOT2020}, FSC2F\cite{VOT2020}, CSRDCF\cite{CSRDCF} and SiamFC\cite{SiamFC} trackers on this dataset. The comparison results are shown in Table \ref{tab:vot2020}. Our ALT tracker has the best EAO score compared to other trackers. This is because our ALT adopts the same backbone network as the DiMP50 \cite{DiMP} tracker, and the accuracy score is slightly lower than this tracker, which not only indicates the importance of the Tversky loss in the target boundary box estimation but also indicates that our method can select more diverse and representative samples for the deep model training.

\begin{figure*} [!t]
\centering
\centerline{{\includegraphics[width=0.9\textwidth]{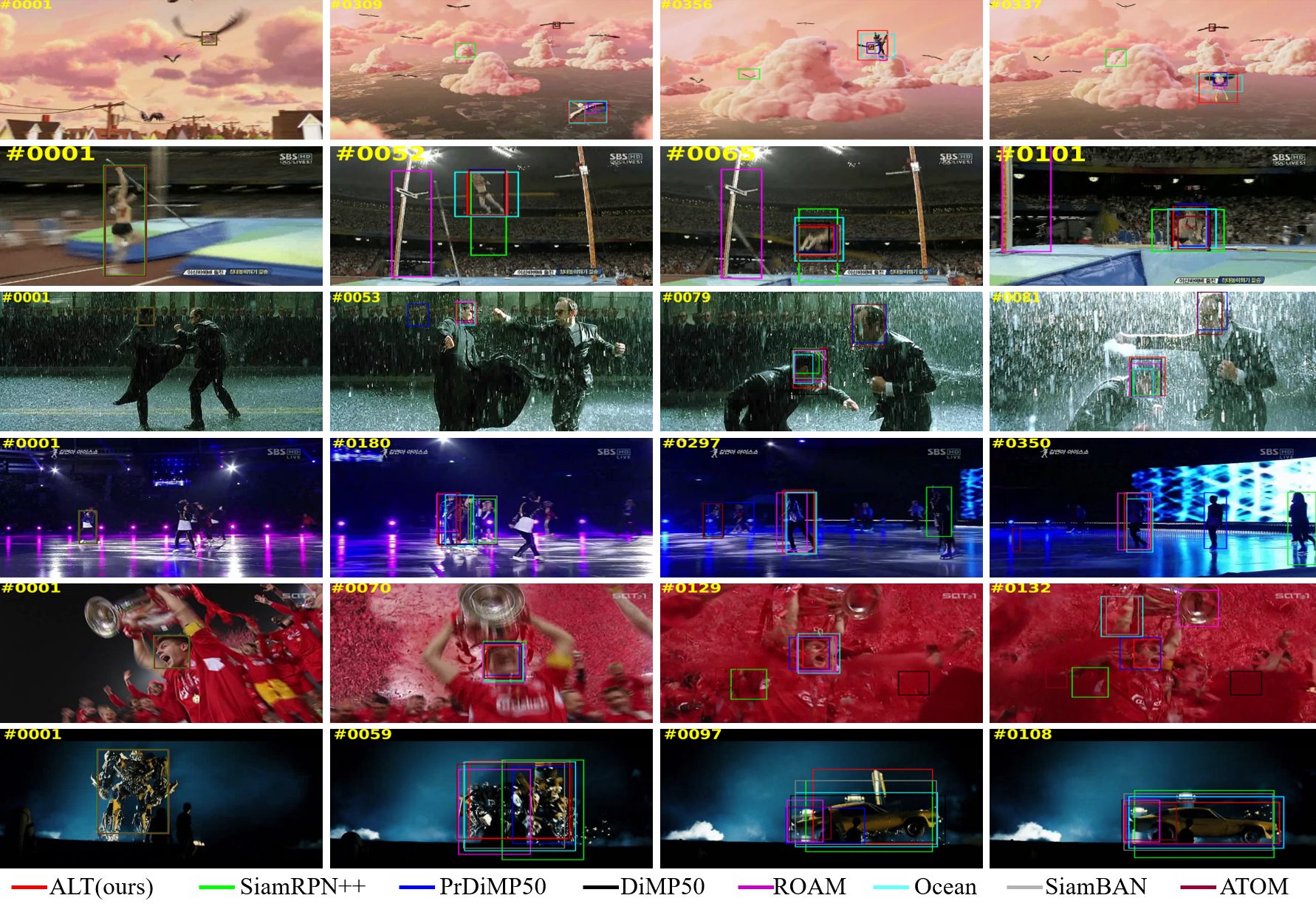}}}
\caption{Qualitative comparison results of our ALT tracker with the SiamRPN++ \cite{ SiamRPN++}, PrDiMP50 \cite{PrDiMP}, DiMP50 \cite{DiMP}, ROAM \cite{ROAM}, Ocean \cite{Ocean}, SiamBAN \cite{SiamBAN} and ATOM \cite{ATOM} trackers (sequences from top to bottom are: bird1, jump, matrix, skating1, soccer and trans).}
\label{fig:qua}
\end{figure*}

 In summary, comparative results on these seven standard benchmark datasets show that our active learning-based ALT tracker has achieved competitive tracking performance relative to these state-of-the-art tracking methods.
It further shows that these training samples, selected based on the active learning method, have the perfect representational ability under a fixed budget.  This can effectively solve the issue of it being prohibitively expensive and time-consuming to label samples in large quantities. It therefore provides a practical direction for the selection of those training samples with strong representational ability.
 Meanwhile, the Tversky loss-based bounding box estimation strategy can also provide some performance gain to the tracker, as it makes the tracker give more attention to the target than the background.

\subsection{Qualitative comparison}
To further visually display the tracking results of the comparison, we present a qualitative comparison of our ALT tracker with seven state-of-the-art trackers, namely SiamRPN++ \cite{ SiamRPN++}, PrDiMP50 \cite{PrDiMP}, DiMP50 \cite{DiMP}, ROAM \cite{ROAM}, Ocean \cite{Ocean}, SiamBAN \cite{SiamBAN} and ATOM \cite{ATOM}.  Fig. \ref{fig:qua} illustrates the comparison results on some challenging sequences. 
Although the SiamRPN++ \cite{ SiamRPN++} tracker trains its network over a large number of training datasets, its tracking performance is unacceptable when the tracking scenarios under deformation (jump) and occlusion (bird1).  By contrast, our ALT tracker uses limited training samples selected via active learning to train the network and achieves accurate tracking results in the same tracking scenarios (jump and bird1).
 The DiMP50 \cite{DiMP} tracker and the PrDiMP50 \cite{PrDiMP}tracker both readily interfere in the scenarios of occlusion, background cluster, and deformation ($e.g.$, bird1, and soccer). Our ALT tracker adopts Tversky loss to obtain more accurate bounding box estimation for the tracking task.
Compared with other trackers, our tracker can achieve more satisfactory tracking results in these complex tracking scenarios.
In summary, compared with other trackers, our tracker shows more accurate tracking results, which indicates that our training samples selected based on active learning are more diverse and representative.

\section{Conclusions} \label{Con}
In this paper, we propose a multi-frame cooperation-based active learning method for training sequences selection to train the deep CNNs model on the tracking task. 
In the proposed tracker, after considering the temporal relation of the target in the sequence, we formulate an active learning method based on multi-frame cooperation for selecting these training samples for labeling, which ensures the diversity of these selected samples, then adopt a nearest neighbor discrimination method based on the average nearest neighbor distance to screen out isolated or low-quality sequences and thereby ensure the representativeness of these selected samples.
Moreover, we adopt the Tversky loss function to improve the tracker's bounding box estimation strategy,  enabling it to obtain a more accurate target state.
Extensive experimental results verify that our ALT tracker, when trained with limited budget samples, can achieve comparable tracking results  when measured against state-of-the-art trackers that require extensive training samples.

\section*{Acknowledgment}
This research was supported by National Natural Science Foundation of China under No. 62172126, by the Shenzhen Research Council under No. JCYJ20210324120202006, by the Special Research project on COVID-19 Prevention and Control of Guangdong Province under No.2020KZDZDX1227. Di Yuan was supported by a scholarship from China Scholarship Council (CSC No. 201906120405).
Dr Xiaojun Chang was partially supported by Australian Research Council (ARC) Discovery Early
Career Researcher Award (DECRA) under grant no. DE190100626.

\begin{bibliographystyle}{IEEEtran}
\begin{bibliography}{IEEEabrv,mybibfile}
\end{bibliography}
\end{bibliographystyle}

\begin{IEEEbiography}[{\includegraphics[width=1in,height=1.25in,clip,keepaspectratio]{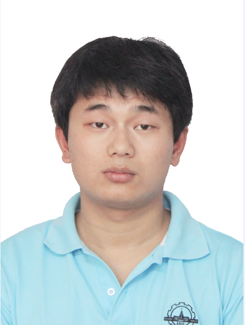}}]
{Di Yuan} received his Ph.D. degree in computer applied technology from Harbin Institute of Technology in 2021. Currently, he is a lecturer at Guangzhou Institute of Technology, Xidian University, Guangzhou, China. He was sponsored by China Scholarship Council as a Visiting Ph.D. student at the Faculty of Information Technology, Monash University Clayton Campus, Australia from 2019 to 2021, working with Prof. Xiaojun Chang. His current research interests include object tracking, machine learning, self-supervised learning, and active learning.
\vspace{-10mm}
\end{IEEEbiography}

\begin{IEEEbiography}[{\includegraphics[width=1in,height=1.25in,clip,keepaspectratio]{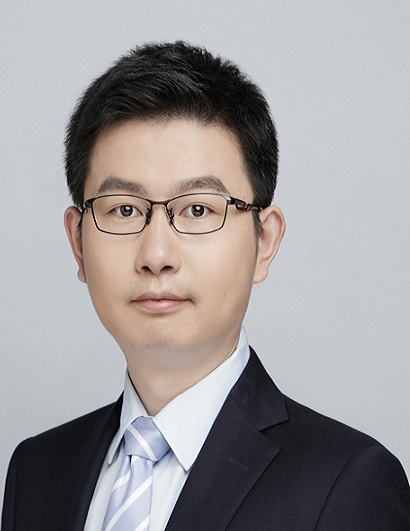}}] {Xiaojun Chang} (Senior Member, IEEE) is a Professor at Faculty of Engineering and Information Technology, University of Technology Sydney.
He was an ARC Discovery Early Career Researcher Award (DECRA) Fellow between 2019-2021. After graduation, he was worked as a Postdoc Research Associate in School of Computer Science, Carnegie Mellon University, a Senior Lecturer in Faculty of Information Technology, Monash University, and an Associate Professor in School of Computing Technologies, RMIT University. He mainly worked on exploring multiple signals for automatic content analysis in unconstrained or surveillance videos and has achieved top performance in various international competitions. 
\vspace{-10mm}
\end{IEEEbiography}

\begin{IEEEbiography}[{\includegraphics[width=1in,height=1.25in,clip,keepaspectratio]{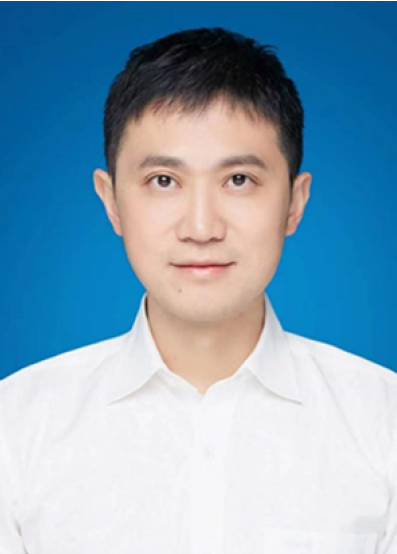}}]
{Yi Yang} (Senior Member, IEEE) received the Ph.D. degree in computer science from Zhejiang University, Hangzhou, China, in 2010. He was a Postdoctoral Researcher with the School of Computer Science, Carnegie Mellon University, Pittsburgh, PA, USA. He is currently a Professor with the University of Technology Sydney, Australia. His current research interests include machine learning and its applications to multimedia content analysis and computer vision, such as multimedia indexing and retrieval, surveillance video analysis, and video content understanding.
\end{IEEEbiography}

\begin{IEEEbiography}[{\includegraphics[width=1in,height=1.25in,clip,keepaspectratio]{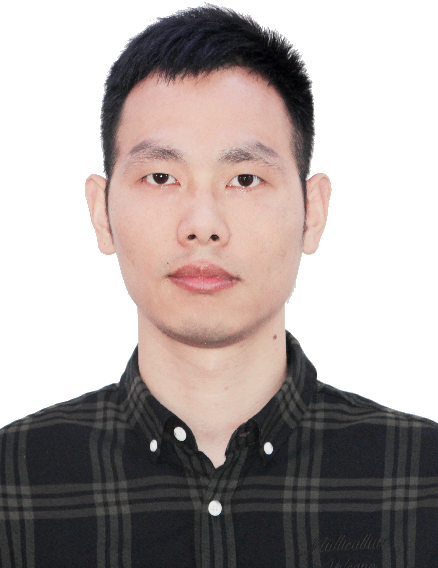}}]
{Qiao Liu} received the PhD degree in computer science from Harbin Institute of Technology in 2021. Currently, he is a lecturer in National Center for Applied Mathematics in Chongqing, Chongqing Normal University. His research interests include thermal infrared object tracking, infrared image processing, and machine learning.
\vspace{-10mm}
\end{IEEEbiography}

\begin{IEEEbiography}[{\includegraphics[width=1in,height=1.25in,clip,keepaspectratio]{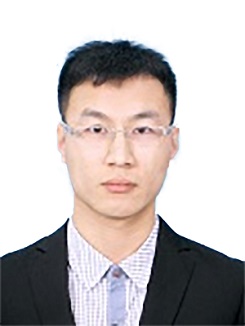}}] {Dehua Wang} received the M.S. degrees in Applied Mathematics from Harbin Institute of Technology, Shenzhen, China in 2017. His current research interests include image processing and machine learning. 
\vspace{-10mm}
\end{IEEEbiography}

\begin{IEEEbiography}[{\includegraphics[width=1in,height=1.25in,clip,keepaspectratio]{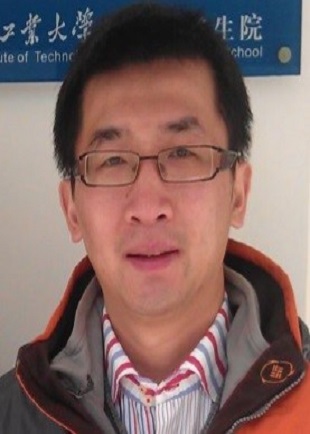}}] {Zhenyu He} (SM'12) received his Ph.D. degree from the Department of Computer Science, Hong Kong Baptist University, Hong Kong, in 2007.  From 2007 to 2009, he worked as a postdoctoral researcher in the department of Computer Science and Engineering, Hong Kong University of Science and Technology. He is currently a full professor in the School of Computer Science and Technology, Harbin Institute of Technology, Shenzhen, China. His research interests include machine learning, computer vision, image processing and pattern recognition. 
\vspace{-10mm}
\end{IEEEbiography}


\begin{thebibliography}{10}
\providecommand{\url}[1]{#1}
\csname url@samestyle\endcsname
\providecommand{\newblock}{\relax}
\providecommand{\bibinfo}[2]{#2}
\providecommand{\BIBentrySTDinterwordspacing}{\spaceskip=0pt\relax}
\providecommand{\BIBentryALTinterwordstretchfactor}{4}
\providecommand{\BIBentryALTinterwordspacing}{\spaceskip=\fontdimen2\font plus
\BIBentryALTinterwordstretchfactor\fontdimen3\font minus
  \fontdimen4\font\relax}
\providecommand{\BIBforeignlanguage}[2]{{%
\expandafter\ifx\csname l@#1\endcsname\relax
\typeout{** WARNING: IEEEtran.bst: No hyphenation pattern has been}%
\typeout{** loaded for the language `#1'. Using the pattern for}%
\typeout{** the default language instead.}%
\else
\language=\csname l@#1\endcsname
\fi
#2}}
\providecommand{\BIBdecl}{\relax}
\BIBdecl

\bibitem{OTB100}
Y.~Wu, J.~Lim, and M.-H. Yang, ``Object tracking benchmark,'' \emph{IEEE
  Transactions on Pattern Analysis and Machine Intelligence}, vol.~37, no.~9,
  pp. 1834--1848, 2015.

\bibitem{UAV123}
M.~Mueller, N.~Smith, and B.~Ghanem, ``A benchmark and simulator for {UAV}
  tracking,'' in \emph{ECCV}, 2016, pp. 445--461.

\bibitem{CFTM}
F.~Liu, C.~Gong, X.~Huang, T.~Zhou, J.~Yang, and D.~Tao, ``Robust visual
  tracking revisited: From correlation filter to template matching,''
  \emph{IEEE Transactions on Image Processing}, vol.~27, no.~6, pp. 2777--2790,
  2018.

\bibitem{CFNet}
J.~Valmadre, L.~Bertinetto, J.~Henriques, A.~Vedaldi, and P.~H.~S. Torr,
  ``End-to-end representation learning for correlation filter based tracking,''
  in \emph{CVPR}, 2017, pp. 2085--2813.

\bibitem{ATOM}
M.~Danelljan, G.~Bhat, F.~S. Khan, and M.~Felsberg, ``A{TOM}: Accurate tracking
  by overlap maximization,'' in \emph{CVPR}, 2019, pp. 4660--4669.

\bibitem{D3S}
A.~Lukezic, J.~Matas, and M.~Kristan, ``D3{S}-a discriminative single shot
  segmentation tracker,'' in \emph{CVPR}, 2020, pp. 7133--7142.

\bibitem{CCRT}
S.~Ge, C.~Zhang, S.~Li, D.~Zeng, and D.~Tao, ``Cascaded correlation refinement
  for robust deep tracking,'' \emph{IEEE Transactions on Neural Networks and
  Learning Systems}, vol.~32, no.~3, pp. 1276--1288, 2020.

\bibitem{RenXCHLCW21}
P.~Ren, Y.~Xiao, X.~Chang, P.~Huang, Z.~Li, X.~Chen, and X.~Wang, ``A
  comprehensive survey of neural architecture search: Challenges and
  solutions,'' \emph{{ACM} Comput. Surv.}, vol.~54, no.~4, pp. 76:1--76:34,
  2021.

\bibitem{MMTV}
X.~Mei, Z.~Hong, D.~Prokhorov, and D.~Tao, ``Robust multitask multiview
  tracking in videos,'' \emph{IEEE Transactions on Neural Networks and Learning
  Systems}, vol.~26, no.~11, pp. 2874--2890, 2015.

\bibitem{NONT}
B.~Ma, H.~Hu, J.~Shen, Y.~Zhang, L.~Shao, and F.~Porikli, ``Robust object
  tracking by nonlinear learning,'' \emph{IEEE Transactions on Neural Networks
  and Learning Systems}, vol.~29, no.~10, pp. 4769--4781, 2017.

\bibitem{DSTRT}
Y.~Zheng, H.~Song, K.~Zhang, J.~Fan, and X.~Liu, ``Dynamically spatiotemporal
  regularized correlation tracking,'' \emph{IEEE Transactions on Neural
  Networks and Learning Systems}, vol.~31, no.~7, pp. 2336--2347, 2019.

\bibitem{SiamFC}
L.~Bertinetto, J.~Valmadre, J.~F. Henriques, A.~Vedaldi, and P.~H.~S. Torr,
  ``Fully-convolutional {S}iamese networks for object tracking,'' in
  \emph{ECCVW}, 2016, pp. 850--865.

\bibitem{DiMP}
G.~Bhat, M.~Danelljan, L.~V. Gool, and R.~Timofte, ``Learning discriminative
  model prediction for tracking,'' in \emph{ICCV}, 2019, pp. 6182--6191.

\bibitem{C-RPN}
H.~Fan and H.~Ling, ``{S}iamese cascaded region proposal networks for real-time
  visual tracking,'' in \emph{CVPR}, 2019, pp. 7952--7961.

\bibitem{TrackingNet}
M.~Muller, A.~Bibi, S.~Giancola, S.~Alsubaihi, and B.~Ghanem, ``Tracking{N}et:
  {A} large-scale dataset and benchmark for object tracking in the wild,'' in
  \emph{ECCV}, 2018, pp. 300--317.

\bibitem{LaSOT}
H.~Fan, L.~Lin, F.~Yang, P.~Chu, G.~Deng, S.~Yu, H.~Bai, Y.~Xu, C.~Liao, and
  H.~Ling, ``La{SOT}: A high-quality benchmark for large-scale single object
  tracking,'' in \emph{CVPR}, 2019, pp. 5374--5383.

\bibitem{GOT10k}
L.~Huang, X.~Zhao, and K.~Huang, ``G{OT}-10k: A large high-diversity benchmark
  for generic object tracking in the wild,'' \emph{IEEE Transactions on Pattern
  Analysis and Machine Intelligence}, 2019.

\bibitem{SSLCN}
Z.~Ling, X.~Li, W.~Zou, and S.~Guo, ``Semi-supervised learning via
  convolutional neural network for hyperspectral image classification,'' in
  \emph{ICPR}, 2018, pp. 1--6.

\bibitem{AALS}
H.~Yu, X.~Yang, S.~Zheng, and C.~Sun, ``Active learning from imbalanced data: A
  solution of online weighted extreme learning machine,'' \emph{IEEE
  Transactions on Neural Networks and Learning Systems}, vol.~30, no.~4, pp.
  1088--1103, 2019.

\bibitem{ALS}
Y.~Fu, X.~Zhu, and B.~Li, ``A survey on instance selection for active
  learning,'' \emph{Knowledge and Information Systems}, vol.~35, no.~2, pp.
  249--283, 2013.

\bibitem{gal2017deep}
Y.~Gal, R.~Islam, and Z.~Ghahramani, ``Deep bayesian active learning with image
  data,'' in \emph{ICML}, 2017, pp. 1183--1192.

\bibitem{LiuC0Y17}
W.~Liu, X.~Chang, L.~Chen, and Y.~Yang, ``Early active learning with pairwise
  constraint for person re-identification,'' in \emph{Joint European Conference
  on Machine Learning and Knowledge Discovery in Databases}, 2017, pp.
  103--118.

\bibitem{RenXCHLCW20}
P.~Ren, Y.~Xiao, X.~Chang, P.~Huang, Z.~Li, B.~B. Gupta, X.~Chen, and X.~Wang,
  ``A survey of deep active learning,'' \emph{{ACM} Comput. Surv.}, vol.~54,
  no.~9, pp. 180:1--180:40, 2022.

\bibitem{LiuCCPZYH20}
W.~Liu, X.~Chang, L.~Chen, D.~Phung, X.~Zhang, Y.~Yang, and A.~G. Hauptmann,
  ``Pair-based uncertainty and diversity promoting early active learning for
  person re-identification,'' \emph{{ACM} Trans. Intell. Syst. Technol.},
  vol.~11, no.~2, pp. 21:1--21:15, 2020.

\bibitem{YangMNCH15}
Y.~Yang, Z.~Ma, F.~Nie, X.~Chang, and A.~G. Hauptmann, ``Multi-class active
  learning by uncertainty sampling with diversity maximization,'' \emph{Int. J.
  Comput. Vis.}, vol. 113, no.~2, pp. 113--127, 2015.

\bibitem{ChangNWYZZ16}
X.~Chang, F.~Nie, S.~Wang, Y.~Yang, X.~Zhou, and C.~Zhang, ``Compound rank-k
  projections for bilinear analysis,'' \emph{{IEEE} Trans. Neural Networks
  Learn. Syst.}, vol.~27, no.~7, pp. 1502--1513, 2016.

\bibitem{aghdam2019active}
H.~H. Aghdam, A.~Gonzalez-Garcia, J.~v.~d. Weijer, and A.~M. L{\'o}pez,
  ``Active learning for deep detection neural networks,'' in \emph{ICCV}, 2019,
  pp. 3672--3680.

\bibitem{ALIC}
W.~H. Beluch, T.~Genewein, A.~N{\"u}rnberger, and J.~M. K{\"o}hler, ``The power
  of ensembles for active learning in image classification,'' in \emph{CVPR},
  2018, pp. 9368--9377.

\bibitem{pinsler2019bayesian}
R.~Pinsler, J.~Gordon, E.~Nalisnick, and J.~M. Hern{\'a}ndez-Lobato, ``Bayesian
  batch active learning as sparse subset approximation,'' in \emph{NIPS}, 2019,
  pp. 6359--6370.

\bibitem{yoo2019learning}
D.~Yoo and I.~S. Kweon, ``Learning loss for active learning,'' in \emph{CVPR},
  2019, pp. 93--102.

\bibitem{BSegm}
X.~Shu, Y.~Yang, and B.~Wu, ``A neighbor level set framework minimized with the
  split {B}regman method for medical image segmentation,'' \emph{Signal
  Processing}, p. 108293, 2021.

\bibitem{ALSegm}
Y.~Siddiqui, J.~Valentin, and M.~Nie{\ss}ner, ``Viewal: Active learning with
  viewpoint entropy for semantic segmentation,'' in \emph{CVPR}, 2020, pp.
  9433--9443.

\bibitem{SNNL}
X.~Shu, Y.~Yang, and B.~Wu, ``Adaptive segmentation model for liver ct images
  based on neural network and level set method,'' \emph{Neurocomputing}, vol.
  453, pp. 438--452, 2021.

\bibitem{C-COT}
M.~Danelljan, A.~Robinson, F.~S. Khan, and M.~Felsberg, ``Beyond correlation
  filters: Learning continuous convolution operators for visual tracking,'' in
  \emph{ECCV}, 2016, pp. 472--488.

\bibitem{ECO}
M.~Danelljan, G.~Bhat, F.~Shahbaz~Khan, and M.~Felsberg, ``E{CO}: Efficient
  convolution operators for tracking,'' in \emph{CVPR}, 2017, pp. 6638--6646.

\bibitem{DMLST}
Q.~Liu, X.~Li, Z.~He, N.~Fan, D.~Yuan, and H.~Wang, ``Learning deep multi-level
  similarity for thermal infrared object tracking,'' \emph{IEEE Transactions on
  Multimedia}, vol.~23, pp. 2114--2126, 2020.

\bibitem{Tversky}
S.~S.~M. Salehi, D.~Erdogmus, and A.~Gholipour, ``Tversky loss function for
  image segmentation using 3d fully convolutional deep networks,'' in
  \emph{International Workshop on Machine Learning in Medical Imaging}, 2017,
  pp. 379--387.

\bibitem{ROAM}
T.~Yang, P.~Xu, R.~Hu, H.~Chai, and A.~B. Chan, ``R{OAM}: Recurrently
  optimizing tracking model,'' in \emph{CVPR}, 2020, pp. 6718--6727.

\bibitem{Ocean}
Z.~Zhang and H.~Peng, ``Ocean: Object-aware anchor-free tracking,'' in
  \emph{ECCV}, 2020.

\bibitem{SiamBAN}
Z.~Chen, B.~Zhong, G.~Li, S.~Zhang, and R.~Ji, ``Siamese box adaptive network
  for visual tracking,'' in \emph{CVPR}, 2020, pp. 6668--6677.

\bibitem{VOT2019}
M.~Kristan, J.~Matas, A.~Leonardis, M.~Felsberg, R.~Pflugfelder,
  \emph{et~al.}, ``The seventh visual object tracking {VOT}2019 challenge
  results,'' in \emph{ICCV Workshops}, 2019.

\bibitem{VOT2020}
M.~Kristan, A.~Leonardis, J.~Matas, M.~Felsberg, R.~Pflugfelder, J.-K.
  K{\"a}m{\"a}r{\"a}inen, M.~Danelljan, L.~{\v{C}}. Zajc,
  A.~Luke{\v{z}}i{\v{c}}, O.~Drbohlav \emph{et~al.}, ``The eighth visual object
  tracking {VOT2020} challenge results,'' in \emph{ECCV}, 2020, pp. 547--601.

\bibitem{SiamHA}
J.~Shen, X.~Tang, X.~Dong, and L.~Shao, ``Visual object tracking by
  hierarchical attention {S}iamese network,'' \emph{IEEE Transactions on
  Cybernetics}, vol.~50, no.~7, pp. 3068--3080, 2019.

\bibitem{SINT}
R.~Tao, E.~Gavves, and A.~W.~M. Smeulders, ``Siamese instance search for
  tracking,'' in \emph{CVPR}, 2016, pp. 1420--1429.

\bibitem{SSDCT}
D.~Yuan, X.~Chang, P.-Y. Huang, Q.~Liu, and Z.~He, ``Self-supervised deep
  correlation tracking,'' \emph{IEEE Transactions on Image Processing},
  vol.~30, pp. 976--985, 2021.

\bibitem{SiamDW}
Z.~Zhang and H.~Peng, ``Deeper and wider {S}iamese networks for real-time
  visual tracking,'' in \emph{CVPR}, 2019, pp. 4591--4600.

\bibitem{SiamFAFM}
W.~Han, X.~Dong, F.~S. Khan, L.~Shao, and J.~Shen, ``Learning to fuse
  asymmetric feature maps in {S}iamese trackers,'' in \emph{CVPR}, 2021, pp.
  16\,570--16\,580.

\bibitem{LuoCNYHZ18}
M.~Luo, X.~Chang, L.~Nie, Y.~Yang, A.~G. Hauptmann, and Q.~Zheng, ``An adaptive
  semisupervised feature analysis for video semantic recognition,''
  \emph{{IEEE} Trans. Cybern.}, vol.~48, no.~2, pp. 648--660, 2018.

\bibitem{SiamRPN++}
B.~Li, W.~Wu, Q.~Wang, F.~Zhang, J.~Xing, and J.~Yan, ``Siam{RPN}++: Evolution
  of {S}iamese visual tracking with very deep networks,'' in \emph{CVPR}, 2019,
  pp. 4282--4291.

\bibitem{SiamAttn}
Y.~Yu, Y.~Xiong, W.~Huang, and M.~R. Scott, ``Deformable {S}iamese attention
  networks for visual object tracking,'' in \emph{CVPR}, 2020, pp. 6728--6737.

\bibitem{SiamRCNN}
P.~Voigtlaender, J.~Luiten, P.~H. Torr, and B.~Leibe, ``Siam {R-CNN}: Visual
  tracking by re-detection,'' in \emph{CVPR}, 2020, pp. 6578--6588.

\bibitem{SiamCAR}
D.~Guo, J.~Wang, Y.~Cui, Z.~Wang, and S.~Chen, ``Siam{CAR}: {S}iamese fully
  convolutional classification and regression for visual tracking,'' in
  \emph{CVPR}, 2020, pp. 6269--6277.

\bibitem{ALSST}
F.~Zhao, T.~Zhang, Y.~Wu, M.~Tang, and J.~Wang, ``Antidecay {LSTM} for
  {S}iamese tracking with adversarial learning,'' \emph{IEEE Transactions on
  Neural Networks and Learning Systems}, vol.~32, no.~10, pp. 4475--4489, 2021.

\bibitem{DHODRL}
X.~Dong, J.~Shen, W.~Wang, L.~Shao, H.~Ling, and F.~Porikli, ``Dynamical
  hyperparameter optimization via deep reinforcement learning in tracking,''
  \emph{IEEE Transactions on Pattern Analysis and Machine Intelligence},
  vol.~43, no.~5, pp. 1515--1529, 2021.

\bibitem{DOTSL}
X.~Lu, C.~Ma, J.~Shen, X.~Yang, I.~Reid, and M.-H. Yang, ``Deep object tracking
  with shrinkage loss,'' \emph{IEEE Transactions on Pattern Analysis and
  Machine Intelligence}, 2020.

\bibitem{PrDiMP}
M.~Danelljan, L.~V. Gool, and R.~Timofte, ``Probabilistic regression for visual
  tracking,'' in \emph{CVPR}, 2020, pp. 7183--7192.

\bibitem{sener2018active}
O.~Sener and S.~Savarese, ``Active learning for convolutional neural networks:
  A core-set approach,'' in \emph{ICLR}, 2018, pp. 1--13.

\bibitem{huang2014active}
S.-J. Huang, R.~Jin, and Z.-H. Zhou, ``Active learning by querying informative
  and representative examples,'' \emph{IEEE Transactions on Pattern Analysis
  and Machine Intelligence}, vol.~10, no.~36, pp. 1936--1949, 2014.

\bibitem{fu2018scalable}
W.~Fu, M.~Wang, S.~Hao, and X.~Wu, ``Scalable active learning by approximated
  error reduction,'' in \emph{KDD}, 2018, pp. 1396--1405.

\bibitem{yan2020active}
Y.~Yan, S.-J. Huang, S.~Chen, M.~Liao, and J.~Xu, ``Active learning with query
  generation for cost-effective text classification,'' in \emph{AAAI}, 2020,
  pp. 6583--6590.

\bibitem{roy2018deep}
S.~Roy, A.~Unmesh, and V.~P. Namboodiri, ``Deep active learning for object
  detection,'' in \emph{BMVC}, 2018, pp. 91(1--12).

\bibitem{bengar2019temporal}
J.~Z. Bengar, A.~Gonzalez-Garcia, G.~Villalonga, B.~Raducanu, H.~H. Aghdam,
  M.~Mozerov, A.~M. Lopez, and J.~van~de Weijer, ``Temporal coherence for
  active learning in videos,'' in \emph{ICCVW}, 2019, pp. 914--923.

\bibitem{GFSDCF}
T.~Xu, Z.-H. Feng, X.-J. Wu, and J.~Kittler, ``Joint group feature selection
  and discriminative filter learning for robust visual object tracking,'' in
  \emph{ICCV}, 2019, pp. 7950--7960.

\bibitem{ZhouCSSYN20}
R.~Zhou, X.~Chang, L.~Shi, Y.~Shen, Y.~Yang, and F.~Nie, ``Person
  reidentification via multi-feature fusion with adaptive graph learning,''
  \emph{{IEEE} Trans. Neural Networks Learn. Syst.}, vol.~31, no.~5, pp.
  1592--1601, 2020.

\bibitem{PointNet++}
C.~R. Qi, L.~Yi, H.~Su, and L.~J. Guibas, ``Point{N}et++: Deep hierarchical
  feature learning on point sets in a metric space,'' in \emph{NIPS}, 2017, pp.
  7095(1--10).

\bibitem{COCO}
T.-Y. Lin, M.~Maire, S.~Belongie, J.~Hays, P.~Perona, D.~Ramanan,
  P.~Doll{\'a}r, and C.~L. Zitnick, ``Microsoft {COCO}: Common objects in
  context,'' in \emph{ECCV}, 2014, pp. 740--755.

\bibitem{GradNet}
P.~Li, B.~Chen, W.~Ouyang, D.~Wang, X.~Yang, and H.~Lu, ``Grad{N}et:
  Gradient-guided network for visual object tracking,'' in \emph{ICCV}, 2019,
  pp. 6162--6171.

\bibitem{GCT}
J.~Gao, T.~Zhang, and C.~Xu, ``Graph convolutional tracking,'' in \emph{CVPR},
  2019, pp. 4649--4659.

\bibitem{ARCF}
Z.~Huang, C.~Fu, Y.~Li, F.~Lin, and P.~Lu, ``Learning aberrance repressed
  correlation filters for real-time {UAV} tracking,'' in \emph{ICCV}, 2019, pp.
  2891--2900.

\bibitem{UDT}
N.~Wang, Y.~Song, C.~Ma, W.~Zhou, W.~Liu, and H.~Li, ``Unsupervised deep
  tracking,'' in \emph{CVPR}, 2019, pp. 1308--1317.

\bibitem{DCFST}
L.~Zheng, M.~Tang, Y.~Chen, J.~Wang, and H.~Lu, ``Learning feature embeddings
  for discriminant model based tracking-supplementary material,'' in
  \emph{ECCV}, 2020.

\bibitem{PGNet}
B.~Liao, C.~Wang, Y.~Wang, Y.~Wang, and J.~Yin, ``P{G}-{N}et: Pixel to global
  matching network for visual tracking,'' in \emph{ECCV}, 2020.

\bibitem{CGACD}
F.~Du, P.~Liu, W.~Zhao, and X.~Tang, ``Correlation-guided attention for corner
  detection based visual tracking,'' in \emph{CVPR}, 2020, pp. 6836--6845.

\bibitem{CLNet}
X.~Dong, J.~Shen, L.~Shao, and F.~Porikli, ``C{LN}et: A compact latent network
  for fast adjusting {S}iamese trackers,'' in \emph{ECCV}, 2020.

\bibitem{LDES}
L.~Yang, J.~Zhu, W.~Song, Z.~Wang, H.~Liu, and S.~C.~H. Hoi, ``Robust
  estimation of similarity transformation for visual object tracking with
  correlation filters,'' in \emph{AAAI}, 2019, pp. 8666--8673.

\bibitem{SPM}
G.~Wang, C.~Luo, Z.~Xiong, and W.~Zeng, ``S{PM}-{T}racker: Series-parallel
  matching for real-time visual object tracking,'' in \emph{CVPR}, 2019, pp.
  3643--3652.

\bibitem{UpdateNet}
L.~Zhang, A.~Gonzalez-Garcia, J.~v.~d. Weijer, M.~Danelljan, and F.~S. Khan,
  ``Learning the model update for {S}iamese trackers,'' in \emph{ICCV}, 2019,
  pp. 4010--4019.

\bibitem{MAML}
G.~Wang, C.~Luo, X.~Sun, Z.~Xiong, and W.~Zeng, ``Tracking by instance
  detection: A meta-learning approach,'' in \emph{CVPR}, 2020, pp. 6288--6297.

\bibitem{ASRCF}
K.~Dai, D.~Wang, H.~Lu, C.~Sun, and J.~Li, ``Visual tracking via adaptive
  spatially-regularized correlation filters,'' in \emph{CVPR}, 2019, pp.
  4670--4679.

\bibitem{LTMU}
K.~Dai, Y.~Zhang, D.~Wang, J.~Li, H.~Lu, and X.~Yang, ``High-performance
  long-term tracking with meta-updater,'' in \emph{CVPR}, 2020, pp. 6298--6307.

\bibitem{TADT}
X.~Li, C.~Ma, B.~Wu, Z.~He \emph{et~al.}, ``Target-aware deep tracking,'' in
  \emph{CVPR}, 2019, pp. 1369--1378.

\bibitem{MemDTC}
T.~Yang and A.~B. Chan, ``Learning dynamic memory networks for object
  tracking,'' in \emph{ECCV}, 2018, pp. 152--167.

\bibitem{SiamMask}
Q.~Wang, L.~Zhang, L.~Bertinetto, W.~Hu, and P.~H. Torr, ``Fast online object
  tracking and segmentation: A unifying approach,'' in \emph{CVPR}, 2019, pp.
  1328--1338.

\bibitem{UPDT}
G.~Bhat, J.~Johnander, M.~Danelljan, F.~S. Khan, and M.~Felsberg, ``Unveiling
  the power of deep tracking,'' in \emph{ECCV}, 2018, pp. 483--498.

\bibitem{CSRDCF}
A.~Lukezic, T.~Vojir, L.~Cehovin~Zajc, J.~Matas, and M.~Kristan,
  ``Discriminative correlation filter with channel and spatial reliability,''
  in \emph{CVPR}, 2017, pp. 6309--6318.

\end{thebibliography}
\end{document}